\documentclass[11pt]{article}

\usepackage[letterpaper,margin=1in]{geometry}
\usepackage[T1]{fontenc}
\usepackage{lmodern}
\usepackage{microtype}
\usepackage[round]{natbib}
\usepackage{amsmath,amssymb,amsthm,mathtools,booktabs,graphicx}

\newtheorem{theorem}{Theorem}

\newtheorem{proposition}[theorem]{Proposition}

\newtheorem{corollary}[theorem]{Corollary}

\theoremstyle{remark}

\newtheorem{remark}[theorem]{Remark}

\DeclareMathOperator{\Cov}{Cov}

\DeclareMathOperator{\Var}{Var}

\DeclareMathOperator{\KL}{KL}

\DeclareMathOperator{\tr}{tr}

\newcommand{\E}{\mathbb{E}}

\newcommand{\R}{\mathbb{R}}

\newcommand{\eps}{\varepsilon}

\newcommand{\dd}{\mathrm{d}}

\usepackage{url}
\usepackage[colorlinks=true,linkcolor=blue,citecolor=blue,urlcolor=blue]{hyperref}
\hypersetup{
  pdftitle={Tempering Does Not Commute with Diffusion: Posterior-Covariance Geometry of Noisy Scores},
  pdfauthor={Keshav Goyal, Suraj Yadav}
}

\title{Tempering Does Not Commute with Diffusion:\\
Posterior-Covariance Geometry of Noisy Scores}

\author{
  Keshav Goyal$^{1,*,\dagger}$ \qquad Suraj Yadav$^{2,*}$\\[8pt]
  {\small $^{1}$Stem Cell Institute, University of Cambridge, United Kingdom}\\
  {\small $^{2}$Hertie Institute for AI in Brain Health, Eberhard-Karls-Universit\"at T\"ubingen, Germany}\\[4pt]
  {\small\texttt{kg603@cam.ac.uk}, \texttt{suraj-prakash.yadav@uni-tuebingen.de}}\\[6pt]
  {\footnotesize $^{*}$Equal contribution. \quad $^{\dagger}$Corresponding author: \texttt{kg603@cam.ac.uk}}
}
\date{}

\begin{document}

\maketitle

\begin{abstract}
Power tempering rescales the score of a clean distribution, but this relationship generally breaks after diffusion: tempering before noising does not simply rescale the noisy score. We characterize this discrepancy for global power tempering and show that it is governed by the geometry of the denoising posterior. The leading local effect is a posterior-covariance transformation of the noisy score, making the propagated temperature generally anisotropic; a residual captures spatial covariance mismatch and higher-order posterior structure.
This characterization is exact for Gaussian distributions, gives the leading low-noise correction for smooth positive densities, and reveals potentially large non-Gaussian corrections for multimodal posteriors at intermediate noise. We extend the analysis to finite-strength tempering using conditional cumulants and recover known Gaussian temporal-rescaling and time-shifting characterizations of temperature sampling. Experiments across controlled and learned-diffusion settings support the theory. Finally, the same propagated field governs the local DSM perturbation induced by power-tempered synthetic data, giving a diffusion-specific interpretation of negative extrapolation.
\end{abstract}

\section{Introduction}
\label{sec:introduction}
Diffusion models represent a data distribution indirectly, through the family of noisy marginals generated by a forward diffusion process. This distinction becomes important whenever the clean distribution itself is modified. Consider global power tempering,
\begin{equation}
q_{\gamma,0}(x) \propto p_0(x)^\gamma.
\label{eq:intro_clean_power}
\end{equation}
Equivalently, $\gamma$ acts as an inverse temperature: $\gamma>1$ sharpens the distribution, while $\gamma<1$ flattens it. Temperature transformations are widely used to control the concentration and diversity of generative distributions \citep{shih2023longhorizon}, to anneal Boltzmann targets \citep{akhoundsadegh2025pita}, and more recently to flatten concentrated model distributions in generative unlearning \citep{block2026temper}. At the clean level, the transformation is elementary:
\begin{equation*}
\nabla_x \log q_{\gamma,0}(x) = \gamma \nabla_x \log p_0(x).
\end{equation*}
It is therefore tempting to implement temperature control in a diffusion model by applying the same scalar rescaling to its score at every noise level. The difficulty is that a diffusion model does not represent $p_0$ directly. At diffusion time $t$, it represents a noised marginal $p_t$, and in general
\begin{equation}
\mathrm{Diffuse}\!\left(p_0^\gamma\right) \neq \left(\mathrm{Diffuse}(p_0)\right)^\gamma.
\label{eq:intro_noncommute}
\end{equation}
A scalar transformation of the clean density therefore need not remain a scalar transformation after passing through the diffusion channel, a broader endpoint-to-path mismatch also encountered in recent work on diffusion guidance \citep{galashov2026guidance}. Recent work on temperature sampling makes this distinction concrete. Temporal Score Rescaling (TSR) develops time-dependent score rescaling for local temperature control and obtains an exact characterization in the isotropic Gaussian setting \citep{xu2026tsr}. Li et al.\ explicitly consider global target $p_0^\gamma$ and show that direct noisy-score scaling can alter within-mode variance, motivating variance-corrective time shifting \citep{li2026diversify}. These developments suggest a more general question: if clean distribution is power tempered and then diffused, what transformation of the noisy score actually results, and what determines its geometry?

We answer this question through the denoising posterior. Consider the Gaussian forward channel
\begin{equation*}
X_t = a_t X_0 + \sigma_t Z, \qquad Z \sim \mathcal{N}(0,I),
\end{equation*}
and let
\begin{equation*}
s_t(x) = \nabla_x \log p_t(x)
\end{equation*}
denote the score of the original noisy marginal. We study an infinitesimal clean temperature perturbation around $\gamma=1$ and denote by $h_t$ the corresponding change in the noisy score obtained by first tempering the clean distribution and then passing it through the forward diffusion channel.

If the already-noised marginal $p_t$ were tempered directly, its infinitesimal response would simply be $s_t$. Clean tempering produces a different field. We show that their discrepancy admits the exact representation
\begin{equation}
h_t(x)-s_t(x) = \nabla_x G_t(x),
\label{eq:intro_kl}
\end{equation}
where $G_t(x)$ is a Kullback--Leibler divergence involving the denoising posterior $p(X_0\mid X_t=x)$. Thus, the failure of tempering and diffusion to commute is controlled by how the noisy observation changes the distribution of its possible clean origins.

Resolving this posterior information quantity reveals the central geometric structure. Define
\begin{equation}
B_t(x) = I + \sigma_t^2 \nabla_x s_t(x)
= \frac{a_t^2}{\sigma_t^2}\operatorname{Cov}(X_0\mid X_t=x).
\label{eq:intro_B}
\end{equation}
These are standard Gaussian denoising/Tweedie identities \citep{efron2011tweedie,meng2021higher,manor2024posterior}. In particular, recent diffusion methods explicitly exploit the posterior-covariance/score-curvature relation for covariance estimation and conditional generation \citep{peng2024posteriorcov,rozet2024diffusionpriors,rissanen2025freehunch}. We show that the clean-tempering response can be decomposed as
\begin{equation}
h_t(x) = B_t(x)s_t(x) + r_t(x),
\label{eq:intro_main_decomposition}
\end{equation}
where $r_t$ combines spatial variation in covariance mismatch with posterior shape beyond first two moments.

Equation~\eqref{eq:intro_main_decomposition} gives a geometric interpretation of how temperature passes through diffusion. Although the clean transformation uses a single scalar $\gamma$, the corresponding noisy-score transformation is generally anisotropic. The denoising posterior describes which clean states remain compatible with the observed noisy point. Its covariance therefore measures, direction by direction, how much uncertainty about the clean sample remains after observing $X_t=x$. Directions that remain weakly resolved respond differently to clean tempering from directions that are already tightly determined. The matrix $B_t$ captures the local covariance transformation, while $r_t$ accounts for spatial covariance mismatch and additional non-Gaussian or multimodal posterior structure.

This decomposition yields a natural regime map. For Gaussian distributions, the denoising posterior is Gaussian and the residual vanishes exactly, so posterior covariance completely determines the infinitesimal response at every noise level. For smooth positive densities satisfying local regularity and standard integrability conditions, we prove that the posterior becomes locally Gaussian at a controlled spatial-derivative rate. Uniformly on compact sets at low noise,
\begin{equation}
h_t - B_t s_t = O(\sigma_t^4), \qquad h_t - s_t = O(\sigma_t^2).
\label{eq:intro_low_noise}
\end{equation}
Posterior covariance therefore captures leading correction to naive radial score scaling. The posterior-negentropy gradient is higher order, $\nabla_x\nu_t=O(\sigma_t^6)$, while the leading fourth-order residual arises from spatial variation in covariance mismatch. The leading coefficient of $\nu_t$ agrees with classical Edgeworth/cumulant expansion of negentropy \citep{jones1987projection,comon1994ica}, and a pointwise heat-flow identity provides the compact-uniform $C^1$ control needed for the score residual. At intermediate noise in multimodal distributions, however, multiple clean states may remain plausible for same noisy observation, making residual substantial and limiting purely covariance-based descriptions of clean tempering.

The Gaussian specialization connects our characterization to recent temperature-sampling methods. For isotropic Gaussians, $B_t$ reduces to scalar resolved-variance factor, recovering the temporal scaling underlying TSR \citep{xu2026tsr}; for anisotropic Gaussians, response becomes matrix-valued across covariance eigendirections. We extend the analysis to finite changes in $\gamma$ using conditional cumulants of denoising posterior. For Gaussian targets, the powered distribution remains Gaussian, yielding the exact temporal shift rescaling characterization, although cumulant expansion itself need not truncate. For non-Gaussian targets, expansion systematically describes finite-strength departures from first-order response \citep{li2026diversify}.

As a secondary consequence, we ask how the propagated field enters denoising score matching. Under locally power-tempered synthetic law, DSM responds to the noised perturbation $h_t$, whose population-gradient effect is its parameter-space pullback. With reference marginals and posterior geometry frozen, this pullback is the gradient of an explicit score potential. This yields a conditional interpretation of synthetic self-training and negative extrapolation: sharpening moves locally along ascent of the frozen potential, while extrapolation in the opposite direction moves along descent. We view this as a model of structured synthetic bias rather than practical samplers in general.

Our experiments follow the theoretical hierarchy: anisotropic Gaussians verify the exact covariance geometry, smooth non-Gaussian targets the low-noise order separation, multimodal targets the posterior residual, and finite-temperature experiments the conditional-cumulant hierarchy. We also verify the distribution-to-training bridge at population level and in a controlled learned Gaussian diffusion model, where sampler distortion is approximately power-like and negative extrapolation illustrates the resulting geometry. Additional non-Gaussian learned-model results appear in Appendix~\ref{app:mnistpca}.

\section{Related Work}
\label{sec:related_work}
Our posterior characterization connects temperature sampling \citep{xu2026tsr,li2026diversify}, endpoint-to-path mismatch in guidance \citep{galashov2026guidance}, Gaussian denoising identities \citep{efron2011tweedie,meng2021higher,manor2024posterior}, and synthetic self-training \citep{alemohammad2026neon}. The distinction is that we characterize the noisy-score response induced by powering the clean law, separating posterior covariance from spatial covariance mismatch and non-Gaussian posterior shape. Appendix~\ref{app:related_work} gives the full comparison with these literatures.

\section{Clean Tempering Through the Diffusion Channel}

\label{sec:propagation}

We now formalize the propagation problem introduced in Section~\ref{sec:introduction}. Let $p_0$ be a positive $C^2$ density on $\R^d$ and consider the Gaussian forward channel
\begin{equation*}
X_t=a_tX_0+\sigma_t Z, \qquad Z\sim\mathcal N(0,I),
\end{equation*}
with noisy marginal density $p_t$ and score $s_t(x)=\nabla_x\log p_t(x)$. We write
\begin{equation*}
\pi_{t,x}(\dd x_0):=p(X_0\in \dd x_0\mid X_t=x)
\end{equation*}
for the denoising posterior. Throughout, differentiation and interchange operations are assumed justified by the regularity conditions collected in Appendix~\ref{app:regularity}.

\subsection{Propagation of a clean tilt}

A useful starting point is an arbitrary infinitesimal exponential tilt of the clean law. For a measurable observable $f$, define
\begin{equation*}
q^f_{\eps,0}(x)=\frac{p_0(x)e^{\eps f(x)}}{\E_{p_0}e^{\eps f(X_0)}}.
\end{equation*}
\begin{proposition}[Tilt propagation]

\label{prop:tilt}

At $\eps=0$,
\begin{align}
\partial_\eps\log q^f_{\eps,t}(x)
&=\E[f(X_0)\mid X_t=x]-\E f(X_0),\label{eq:tiltdensity}\\
\partial_\eps\nabla_x\log q^f_{\eps,t}(x)
&=\nabla_x\E[f(X_0)\mid X_t=x]\nonumber\\
&=\frac{a_t}{\sigma_t^2}\Cov(f(X_0),X_0\mid X_t=x).
\label{eq:tiltscore}
\end{align}
\end{proposition}

\begin{proof}

By Bayes' rule, the noised tilted density can be written as
\begin{equation*}
q^f_{\eps,t}(x)=p_t(x)\,\frac{\E[e^{\eps f(X_0)}\mid X_t=x]}{\E[e^{\eps f(X_0)}]}.
\end{equation*}
Differentiating its logarithm at $\eps=0$ gives~\eqref{eq:tiltdensity}; differentiating once more in $x$ gives the first equality in~\eqref{eq:tiltscore}. For the Gaussian channel,
\begin{equation*}
\nabla_x\log \pi_{t,x}(X_0)=\frac{a_t}{\sigma_t^2}\left(X_0-\E[X_0\mid X_t=x]\right).
\end{equation*}
Therefore differentiating the posterior expectation yields
\begin{equation*}
\nabla_x\E[f(X_0)\mid X_t=x]=\frac{a_t}{\sigma_t^2}\Cov(f(X_0),X_0\mid X_t=x),
\end{equation*}
which proves~\eqref{eq:tiltscore}.
\end{proof}

The forward channel therefore does not transport a clean perturbation pointwise. It replaces the clean observable by its conditional expectation under the denoising posterior. This is the basic mechanism behind all subsequent results: diffusion converts a transformation specified at the clean endpoint into a posterior-averaged transformation at every noisy level. Global power tempering is the self-tilt $f=\log p_0$. Writing $L(x)=\log p_0(x)$ and
\begin{equation*}
q_{\gamma,0}(x)=\frac{p_0(x)^\gamma}{\int p_0(u)^\gamma\dd u},
\end{equation*}
we define the infinitesimal noisy-score response
\begin{equation*}
h_t(x):=\left.\partial_\gamma\nabla_x\log q_{\gamma,t}(x)\right|_{\gamma=1}.
\end{equation*}
Proposition~\ref{prop:tilt} immediately gives
\begin{equation*}
h_t(x)=\nabla_x\E[L(X_0)\mid X_t=x].
\end{equation*}
Thus the response to a scalar clean temperature is already a posterior-dependent vector field after noising.

\subsection{The posterior information potential}

To compare clean tempering with direct noisy-score scaling, define the normalized Gaussian channel law in clean coordinates,
\begin{equation*}
\Lambda^{\flat}_{t,x}:=\mathcal N\!\left(\frac{x}{a_t},\frac{\sigma_t^2}{a_t^2}I\right).
\end{equation*}
\begin{theorem}[Posterior information characterization]

\label{thm:kl}

For every $t>0$,
\begin{equation}
h_t(x)-s_t(x)=\nabla_xG_t(x), \qquad G_t(x):=\KL(\pi_{t,x}\|\Lambda^{\flat}_{t,x}).
\label{eq:klmain}
\end{equation}
\end{theorem}

A direct power of the already-noised density would have infinitesimal response $s_t$. Theorem~\ref{thm:kl} shows exactly what clean tempering adds: the spatial gradient of a denoising-posterior information quantity. The mismatch is therefore controlled by how the posterior information about possible clean origins varies across noisy space, rather than by a scalar correction depending only on $t$. The proof is given in Appendix~\ref{app:klproof}.

\section{Posterior-Covariance Geometry}

\label{sec:geometry}

The posterior information potential can be resolved into a local covariance contribution and a residual that measures posterior structure beyond second order. The first two Gaussian-denoising identities are
\begin{align}
m_t(x)&:=\E[X_0\mid X_t=x]=\frac{x+\sigma_t^2s_t(x)}{a_t},\label{eq:tweedie1}\\
B_t(x)&:=I+\sigma_t^2\nabla_xs_t(x)
=\frac{a_t^2}{\sigma_t^2}\Cov(X_0\mid X_t=x).
\label{eq:tweedie2}
\end{align}
The matrix $B_t$ is positive semidefinite for exact score and is normalized covariance of denoising posterior.

\begin{theorem}[Posterior-covariance decomposition]

\label{thm:decomp}

Let $\Gamma_{t,x}$ be the Gaussian distribution with the same mean and covariance as $\pi_{t,x}$. Wherever $B_t(x)\succ0$,
\begin{align}
G_t(x)
&=\frac{\sigma_t^2}{2}\|s_t(x)\|^2+\Phi_t(x)+\nu_t(x),
\label{eq:Gdecomp}\\
h_t(x)&=B_t(x)s_t(x)+r_t(x),
\label{eq:hdecomp}\\
r_t(x)&:=\nabla_x\bigl(\Phi_t(x)+\nu_t(x)\bigr),\nonumber
\end{align}
where, for ambient dimension $d$,
\begin{align}
\Phi_t(x)&:=\frac12\left(\operatorname{tr}B_t(x)-d-\log\det B_t(x)\right),\label{eq:phidef}\\
\nu_t(x)&:=\KL(\pi_{t,x}\|\Gamma_{t,x}).
\label{eq:nudef}
\end{align}
\end{theorem}

The decomposition gives central geometric picture of the paper. A single clean temperature becomes matrix-valued transformation $B_ts_t$ after diffusion. The eigenstructure of $B_t$ records how much uncertainty about clean sample remains in each local direction: directions that noisy observation still resolves strongly retain a larger response, while directions that have been washed out respond more weakly. The residual $r_t$ has a distinct role. The term $\Phi_t$ records covariance mismatch relative to Gaussian channel, while $\nu_t$ measures posterior shape information beyond any Gaussian with same first two moments. Thus $r_t$ is residual of transformation $B_ts_t$: its covariance-mismatch component depends on spatial variation in $B_t$, while its negentropy component depends on posterior structure beyond first two moments. Proof in Appendix~\ref{app:decompproof}.

The covariance-mismatch component also admits a temporal-consistency interpretation through the score Fokker--Planck equation; we give the corresponding identity in Appendix~\ref{app:fp}.

\paragraph{Matrix-free form of the covariance response.}

Although $B_t(x)$ has the interpretation of a normalized denoising-posterior covariance, the action that appears in the tempering response does not require explicitly constructing this covariance matrix. From the definition of $B_t$,
\[
B_t(x)s_t(x)
=
s_t(x)
+
\sigma_t^2
\bigl(\nabla_x s_t(x)\bigr)s_t(x).
\]
Hence the leading anisotropic correction requires only the action of the score Jacobian on score vector, rather than estimation or storage of full $d\times d$ posterior covariance matrix. Moreover, for an exact score field $s_t=\nabla_x\log p_t$, the score Jacobian is symmetric, and
\[
\bigl(\nabla_x s_t(x)\bigr)s_t(x)
=
\frac{1}{2}\nabla_x\|s_t(x)\|^2.
\]
Consequently,
\[
B_t(x)s_t(x)
=
s_t(x)
+
\frac{\sigma_t^2}{2}\nabla_x\|s_t(x)\|^2.
\]
Thus the posterior-covariance interpretation is not merely descriptive: the action of the leading geometric operator is encoded directly in local differential information of the noisy score field.

\subsection{Regimes of validity}

\begin{corollary}[Gaussian exactness]
\label{cor:gaussian}
If $p_0=\mathcal N(\mu,\Sigma)$, then $r_t\equiv0$ and \begin{equation} h_t=B_ts_t, \qquad B_t=a_t^2\Sigma(a_t^2\Sigma+\sigma_t^2I)^{-1}. \label{eq:gaussianB} \end{equation}
In the eigenbasis of $\Sigma$, direction $i$ is multiplied by \begin{equation} r_i(t)=\frac{a_t^2\lambda_i}{a_t^2\lambda_i+\sigma_t^2}. \label{eq:gaussian_factor} \end{equation}
\end{corollary}
\begin{proof}

For Gaussian $p_0$, the denoising posterior $\pi_{t,x}$ is Gaussian and its covariance is independent of $x$. Hence $\nu_t=0$ and $\Phi_t$ is spatially constant, so $r_t=\nabla(\Phi_t+\nu_t)=0$. The Gaussian conditioning formula gives
\begin{equation*}
\Cov(X_0\mid X_t=x)=\sigma_t^2\Sigma(a_t^2\Sigma+\sigma_t^2I)^{-1},
\end{equation*}
and substituting into~\eqref{eq:tweedie2} yields~\eqref{eq:gaussianB}; diagonalizing $\Sigma$ gives~\eqref{eq:gaussian_factor}.
\end{proof}

For an isotropic Gaussian all $r_i(t)$ coincide and the response reduces to a scalar temporal factor. For an anisotropic Gaussian they differ by eigendirection, making explicit why a scalar clean temperature generally becomes an anisotropic noisy transformation. In this regime the posterior is exactly Gaussian, so covariance is not an approximation: it contains the complete propagation geometry. For strongly log-concave $p_0\propto e^{-U}$ with $\alpha I\preceq\nabla^2U\preceq LI$, posterior covariance also gives the spectral sandwich
\begin{equation}
\frac{a_t^2}{a_t^2+L\sigma_t^2}I\preceq B_t(x)\preceq\frac{a_t^2}{a_t^2+\alpha\sigma_t^2}I,
\label{eq:logconcave_bound}
\end{equation}
proved in Appendix~\ref{app:decompproof}. Related heat-flow log-Hessian bounds and their implications for diffusion geometry are studied by \citet{brigati2025heat}.

\begin{theorem}[Low-noise accuracy]
\label{thm:lownoise}
Let $K$ be compact and let $U$ be an open neighbourhood of $K$ on which $\ell=\log p_0$ is $C^6$, its derivatives through order six are bounded, and $p_0$ is bounded away from zero. Assume that, for some $\eta\in(0,1)$,
\begin{equation}
\int_{\R^d}\bigl(p_0(u)^{1-\eta}+p_0(u)^{1+\eta}\bigr)\,\dd u<\infty,
\label{eq:lownoise_integrability}
\end{equation}
and that $a_t=1+O(\sigma_t^2)$. Then, uniformly on $K$ as $\sigma_t\to0$,
\begin{equation}
h_t-s_t=O(\sigma_t^2),\qquad h_t-B_ts_t=O(\sigma_t^4).
\label{eq:lownoise}
\end{equation}
Moreover, the posterior-shape term satisfies
\begin{equation}
\nu_t(x)=\frac{\sigma_t^6}{12}\|\nabla^3\ell(x)\|_{\mathrm F}^2+O_{C^1}(\sigma_t^8),
\label{eq:lownoise_negentropy}
\end{equation}
so $\nabla_x\nu_t=O(\sigma_t^6)$. Here $O_{C^1}$ controls both the remainder and its first spatial derivatives, and $\|\nabla^3\ell\|_{\mathrm F}^2=\sum_{i,j,k}(\partial_{ijk}\ell)^2$.
\end{theorem}

The theorem derives local posterior Gaussianity from smoothness and integrability. Posterior covariance captures the leading second-order correction to radial score scaling and leaves a fourth-order residual. The value coefficient $1/12$ in~\eqref{eq:lownoise_negentropy} agrees with the classical leading Edgeworth/negentropy coefficient obtained after centering and whitening the posterior \citep{jones1987projection,comon1994ica}. Appendix~\ref{app:lownoise} further derives an exact pointwise evolution equation for posterior negentropy and establishes a compact-uniform $C^1$ remainder, yielding $\nabla\nu_t=O(\sigma_t^6)$. The same analysis shows that the fourth-order covariance residual has leading coefficient $\frac14\nabla|\nabla^2\ell|_{\mathrm F}^2$, arising from $\nabla\Phi_t$, whereas $\nabla\nu_t$ enters only at sixth order. These leading coefficients may vanish at individual points. The complete proof is given in Appendix~\ref{app:lownoise}.

At intermediate noise in a multimodal distribution, the posterior can instead remain uncertain between several distant clean explanations. In that regime, covariance may no longer summarize the relevant ambiguity, and $r_t$ can be comparable to or larger than $B_ts_t$. The residual is therefore a regime diagnostic, rather than merely a term required to close an algebraic identity.

\subsection{Beyond infinitesimal temperature}

The previous results describe the derivative at $\gamma=1$. Finite temperature changes introduce a second, independent notion of locality. Let $\delta=\gamma-1$. Bayes' rule gives the exact identity

\begin{proposition}[Finite power through the channel]

\label{prop:finite}

For every admissible $\gamma$,
\begin{equation}
\nabla_x\log q_{\gamma,t}(x)=s_t(x)+\nabla_x\log\E\!\left[e^{\delta L(X_0)}\mid X_t=x\right].
\label{eq:finite}
\end{equation}
Whenever the conditional cumulant-generating function is analytic,
\begin{align}
\nabla_x\log q_{\gamma,t}
&=s_t+\delta h_t+\frac{\delta^2}{2}\nabla_x\Var(L(X_0)\mid X_t)\nonumber\\
&\quad+O_t(\delta^3).
\label{eq:cumulant}
\end{align}
\end{proposition}

\begin{proof}
Using $p_0^{1+\delta}=p_0e^{\delta L}$ inside the forward channel and conditioning under the reference model gives
\begin{equation*}
q_{\gamma,t}(x)=\frac{p_t(x)}{Z_\gamma}\,\E[e^{\delta L(X_0)}\mid X_t=x].
\end{equation*}
Taking $x$-gradient of the logarithm proves~\eqref{eq:finite}. Expanding conditional log moment-generating function as
\begin{equation*}
\log\E[e^{\delta L}\mid X_t=x]=\sum_{k\ge1}\frac{\delta^k}{k!}\kappa_k(L\mid X_t=x)
\end{equation*}
and differentiating term by term gives~\eqref{eq:cumulant}.
\end{proof}

The conditional cumulants organize finite-strength corrections around the exact first-order field $h_t$. Combining this expansion with Theorem~\ref{thm:decomp} motivates the covariance approximation
\begin{equation}
s_t^{\rm cov}(\gamma):=s_t+(\gamma-1)B_ts_t.
\label{eq:covsampler}
\end{equation}
Under the assumptions of Theorem~\ref{thm:lownoise} and the stated regularity for the conditional cumulant expansion,
\begin{equation}
\nabla_x\log q_{\gamma,t}-s_t^{\rm cov}(\gamma)
=O(|\delta|\sigma_t^4)+O(\delta^2\sigma_t^2)+O_t(\delta^3).
\label{eq:finiteerror}
\end{equation}
The three terms separate the first-order covariance-approximation error, finite-temperature curvature, and higher temperature cumulants. At low noise, the first term is led by spatial covariance mismatch; posterior-negentropy contribution is $O(|\delta|\sigma_t^6)$. The distinction is important: decreasing noise and temperature strength are different asymptotic limits.

\subsection{Gaussian specialization and recovery of temporal shift--rescaling}

For a zero-mean Gaussian $p_0=\mathcal N(0,\Sigma)$ under a variance-preserving parameterization, clean powering gives covariance $\Sigma/\gamma$. The finite-temperature score can therefore be represented exactly by evaluating the original Gaussian score at a shifted noise level and rescaling it. In particular, choosing $t'$ such that
\begin{equation}
\sigma_{t'}^2=\frac{\gamma\sigma_t^2}{a_t^2+\gamma\sigma_t^2}, \qquad a_{t'}^2=\frac{a_t^2}{a_t^2+\gamma\sigma_t^2},
\label{eq:shift_time}
\end{equation}
gives
\begin{equation}
\nabla_x\log q_{\gamma,t}(x)=\frac{\gamma}{a_t^2+\gamma\sigma_t^2}s_{t'}(x).
\label{eq:shiftrescale}
\end{equation}
Thus posterior formulation recovers the known Gaussian time-shift--rescaling structure studied in recent temperature-sampling work \citep{xu2026tsr,li2026diversify}. The Gaussian exactness here follows from closure of Gaussian family under power tempering and Gaussian noising, rather than from finite truncation of the conditional-cumulant expansion. Its role here is as a specialization and consistency check of general posterior characterization. Away from Gaussian posteriors, residual and higher conditional cumulants describe additional structure that  pure variance correction cannot represent. Further details in Appendix~\ref{app:finite}.

\section{A Training-Space Application: Synthetic Self-Training}
\label{sec:synthetic}

The preceding sections characterize how clean power tempering propagates through diffusion channel. As an application, we ask how this propagated field appears when transformed samples are subsequently used for denoising score matching. The section isolates local training-space consequence of power-tempering perturbation developed above, rather than attempting to characterize synthetic-data pipelines in full generality.

Let $\theta_r$ be self-consistent reference checkpoint with score $s_{\theta_r}=s$, and suppose its synthetic sampler produces clean law $q_{1+\delta,0}\propto p_{\theta_r,0}^{1+\delta}$. Training on resulting noised marginals uses population score-space DSM risk
\begin{align*}
R_{\rm syn}(\theta)
&=\frac12\int\rho(t)\\
&\times\E_{q_{1+\delta,t}}\!\left[\|s_\theta(X,t)-\nabla_x\log q_{1+\delta,t}(X)\|^2\right]\dd t.
\end{align*}
\begin{theorem}[Synthetic DSM tangent]

\label{thm:syn}

Under the local power-tempering model,
\begin{align}
\nabla_\theta R_{\rm syn}(\theta_r)
&=-\delta\int\rho(t)\E_{p_{\theta_r,t}}\!\left[(\partial_\theta s_{\theta_r})^\top h_t\right]\dd t\nonumber\\
&\quad+O(\delta^2).
\label{eq:syngrad}
\end{align}
For $\delta=0$, exact population self-sampling has zero systematic first-order gradient.

\end{theorem}

A complete proof is given in Appendix~\ref{app:synthetic}. The synthetic distribution is generated from the frozen reference checkpoint, so differentiation is only through the trained score $s_\theta$; the clean sampler distortion enters through the perturbed noisy-score field $h_t$. To first order, the resulting gradient can be evaluated under the detached reference-model marginals $p_{\theta_r,t}$, yielding the parameter-space tangent seen by synthetic fine-tuning.

\subsection{A frozen covariance potential}

Freeze the reference-model marginals, $B_t$, and $r_t$ at $\theta_r$, and define
\begin{align}
V_{\rm cov}(\theta)
&:=\frac12\int\rho(t)\nonumber\\
&\times\E_{p_{\theta_r,t}}\!\left[s_\theta(X,t)^\top B_t^{\theta_r}(X)s_\theta(X,t)\right]\dd t,
\label{eq:Vcovmain}\\
W_{\rm def}(\theta)
&:=\int\rho(t)\E_{p_{\theta_r,t}}\!\left[s_\theta(X,t)^\top r_t^{\theta_r}(X)\right]\dd t,\nonumber\\
V_{\rm full}(\theta)&:=V_{\rm cov}(\theta)+W_{\rm def}(\theta).\nonumber
\end{align}
Because the measure and reference geometry are frozen,
\begin{align}
\nabla_\theta V_{\rm cov}(\theta_r)
&=\int\rho(t)\E_{p_{\theta_r,t}}[(\partial_\theta s_{\theta_r})^\top B_t s_t]\dd t,\label{eq:gradVcov}\\
\nabla_\theta W_{\rm def}(\theta_r)
&=\int\rho(t)\E_{p_{\theta_r,t}}[(\partial_\theta s_{\theta_r})^\top r_t]\dd t.\nonumber
\end{align}
Hence Theorem~\ref{thm:syn} and $h_t=B_ts_t+r_t$ give
\begin{equation}
\nabla_\theta R_{\rm syn}(\theta_r)
=-\delta\nabla_\theta V_{\rm full}(\theta_r)+O(\delta^2).
\label{eq:synthetic_potential}
\end{equation}
For sharpening ($\delta>0$), gradient descent on the synthetic loss therefore moves locally in the ascent direction of $V_{\rm full}$. For Gaussian model marginals the defect vanishes, so $V_{\rm cov}$ is the exact frozen potential at first order; more generally it is the covariance approximation singled out by Section~\ref{sec:geometry}.

\subsection{Connection to negative extrapolation}
Negative extrapolation methods such as NEON \citep{alemohammad2026neon} form a synthetic-finetuned checkpoint $\theta_s$ and extrapolate away from it, $\theta_{\rm NE}(w)=\theta_r-w(\theta_s-\theta_r)$. Under the local power-tempering model, this reverses the frozen-potential direction induced by synthetic training. We also test the frozen covariance objective $J_{\rm cov}=R_{\rm anchor}+\mu V_{\rm cov}$, with $\mu\geq0$, as a supporting regularization experiment. Appendix~\ref{app:negative_extrapolation} retains the full derivation, matrix-free interpretation, and connection to isotropic score-energy regularization; Appendix~\ref{app:regularization} gives the parameterization and weighting details.

\section{Experiments}

\label{sec:experiments}

The experiments are organized around the theoretical claims of Sections~\ref{sec:propagation}--\ref{sec:geometry}. We first validate posterior characterization across its exact Gaussian, low-noise non-Gaussian, multimodal, and finite-temperature regimes. We then verify that same propagated field controls the population DSM perturbation induced by power-tempered synthetic data. Finally, controlled learned Gaussian model provides one training-space application by testing the predicted synthetic direction and its connection to negative extrapolation.

\subsection{Distribution-level validation}

\label{sec:distval}

\paragraph{Scalar clean temperature becomes anisotropic after diffusion.}

\begin{figure}[t]
\centering
\includegraphics[width=\linewidth]{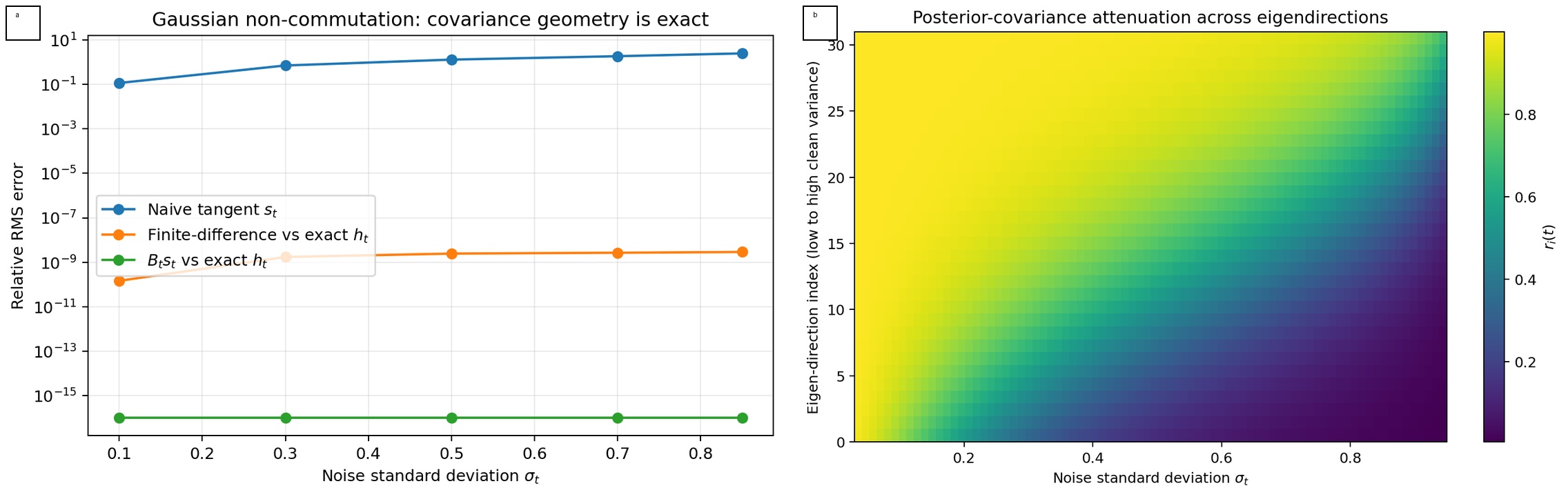}
\caption{\textbf{Exact Gaussian geometry.} Left: the analytic clean-tempering tangent and $B_ts_t$ overlap, whereas naive radial tempering deviates as noise increases. Right: covariance eigendirections are attenuated according to their resolved-variance fractions.}
\label{fig:gaussian}
\end{figure}

On a $32$-dimensional zero-mean Gaussian with covariance eigenvalues log-spaced from $0.05$ to $20$, finite differences of the exact clean-tempered score agree with $B_ts_t$ to numerical precision, while naive radial tempering becomes increasingly inaccurate with noise. At $\sigma_t=0.5$, its relative RMS error is $1.335$. More importantly, the discrepancy is directional: the measured response factors follow
\[
r_i(t)=\frac{a_t^2\lambda_i}{a_t^2\lambda_i+\sigma_t^2},
\]
so low-variance clean directions lose temperature response earlier than high-variance directions (Figure~\ref{fig:gaussian}).

\paragraph{Covariance is a non-Gaussian low-noise principle.}

We next use the smooth one-dimensional density $p_0(x)\propto\exp[-0.25x^2-1.5\log\cosh x]$, for which the noised density and posterior quantities can be evaluated accurately without a neural model. The measured errors scale as
\begin{equation*}
\|h_t-s_t\|\propto\sigma_t^{1.990},
\qquad
\|h_t-B_ts_t\|\propto\sigma_t^{3.933},
\end{equation*}
matching the $O(\sigma_t^2)$ and $O(\sigma_t^4)$ orders in Theorem~\ref{thm:lownoise}. At $\sigma_t=0.10$, the corresponding errors are $1.42\times10^{-2}$ and $6.1\times10^{-5}$. This verifies posterior covariance is leading local correction beyond Gaussian family itself.

\paragraph{The posterior residual identifies the multimodal regime.}

To leave the local Gaussian regime deliberately, we use a symmetric bimodal Gaussian mixture at intermediate noise. Near locations where the noisy observation does not identify a unique clean mode, the exact tangent and $B_ts_t$ separate strongly; near $x=0.12$, for example, $h_t\approx1.91$ while $B_ts_t\approx11.37$. The posterior remains well defined, but covariance alone cannot summarize uncertainty between distant clean explanations (refer to Figure~\ref{fig:regimes}).

\paragraph{Finite-strength corrections follow the conditional-cumulant hierarchy.}

We compare naive scaling, the covariance approximation~\eqref{eq:covsampler}, the exact first-order tangent, and the second conditional-cumulant expansion over a sweep of $\delta=\gamma-1$. On a Gaussian, log--log error slopes are $0.998$ (naive), $1.975$ (covariance/exact first order), and $2.974$ (second cumulant), while the recovered Gaussian shift--rescale formula is at numerical precision. On the GMM, covariance remains first order ($0.972$) because its first-order shape defect is nonzero; using the exact $h_t$ restores a second-order slope ($1.967$), and adding the second cumulant restores a third-order slope ($2.968$). Higher-order temperature terms therefore help only after the correct first-order geometry has been supplied (Figure~\ref{fig:finite}).

Before introducing a learned network, we test Theorem~\ref{thm:syn} in an $8$D anisotropic Gaussian model, where both the exact clean-tempered noisy score and the DSM population gradient can be evaluated analytically without Monte Carlo bias. Across $\delta\in[0.005,0.4]$, the exact synthetic gradient is collinear with the covariance prediction. The relative magnitude error of the first-order covariance approximation grows with $|\delta|$, as expected from the finite-temperature expansion. Under exact self-sampling ($\delta=0$), the empirical gradient has mean near zero and RMS magnitude proportional to $N^{-0.501}$, confirming that the deterministic population signal is induced by the sampler distortion rather than by self-training itself.

\subsection{Controlled learned Gaussian: a training-space application}

\label{sec:learnedapp}

We train a $2$D $\epsilon$-prediction MLP on a Gaussian target with covariance $\begin{psmallmatrix}2&1\\1&2\end{psmallmatrix}$ using $1{,}000$ real samples. A $20$-step reverse sampler with score/noise multiplier $c=1.1$ generates the distorted synthetic samples used by the self-training branch. Because the clean target is Gaussian, the covariance field is exact under the power-tempering model; the remaining empirical question is whether the finite learned sampler produces a sufficiently power-like distortion.

The generated covariance is well fitted by a clean Gaussian power perturbation, with $\gamma_{\rm eff}=1.239\pm0.043$ and covariance-fit error $2.29\%\pm1.17\%$. After subtracting finite-sampler self-training signal at $c=1$, incremental synthetic DSM gradient has cosine $0.9851\pm0.0071$ with the covariance prediction $-\nabla_\theta V_{\rm cov}$ over nine independent seeds. After $250$ synthetic Adam steps, the corresponding checkpoint displacement retains cosine $0.701\pm0.034$ with the initial covariance-predicted synthetic displacement. Thus the local field predicted by the theory remains visible after score approximation, finite sampling, and finite optimization.

The endpoint behavior provides a supporting connection to negative extrapolation. Synthetic fine-tuning degrades FID from $0.0496\pm0.0165$ to $0.2261\pm0.0311$, while extrapolating opposite to the synthetic displacement with $w=0.5$ improves it to $0.0177\pm0.0148$, with improvement over the base in all nine seeds. We also optimize the frozen covariance objective~\eqref{eq:Jcov}; it reaches FID $0.00780\pm0.00589$, compared with $0.0455\pm0.0106$ for a matched continuation of ordinary real-data DSM training. In this Gaussian objective experiment the covariance term is evaluated on noised real samples as a proxy for the detached reference-model measure; the two covariance gradients have cosine $0.9975\pm0.0012$. Table~\ref{tab:gaussian_app} summarizes these endpoint results. Additional Gaussian checks and a non-Gaussian MNIST-PCA extension are reported in Appendix~\ref{app:extraexp}.

\begin{table}[t]

\caption{Controlled learned-Gaussian application.}

\label{tab:gaussian_app}

\begin{center}
\begin{tabular}{lr}
\toprule
Method & FID $\downarrow$ \\
\midrule
Base & $0.0496\pm0.0165$ \\
Synthetic fine-tuning & $0.2261\pm0.0311$ \\
Negative extrapolation, $w=0.5$ & $0.0177\pm0.0148$ \\
Ordinary real-data continuation & $0.0455\pm0.0106$ \\
Frozen covariance objective & $\mathbf{0.00780\pm0.00589}$ \\
\bottomrule
\end{tabular}
\end{center}

\end{table}

\section{Conclusion}

A scalar power applied to a clean density generally becomes an anisotropic, posterior-dependent transformation after diffusion. We characterize this transformation through the denoising posterior, decomposing the infinitesimal noisy-score response into a posterior-covariance component and an explicit residual combining spatial covariance mismatch and non-Gaussian posterior shape. Covariance is exact for Gaussians and captures the leading low-noise correction locally for smooth positive densities under standard integrability conditions. A pointwise heat-flow argument gives compact-uniform $C^1$ control of posterior negentropy; its leading value coefficient agrees with the classical Edgeworth/cumulant negentropy expansion, while the leading fourth-order score residual comes instead from spatial covariance mismatch. At intermediate noise, strongly non-Gaussian or multimodal posterior structure can require substantial corrections. Conditional cumulants extend the same picture to finite temperature strength.

Our experiments validate this hierarchy across Gaussian, non-Gaussian, multimodal, finite-temperature, and population-DSM settings. As one application, the same propagated field gives a local interpretation of power-tempered synthetic self-training and negative extrapolation, which we verify in a controlled learned Gaussian model.

The present analysis is limited to global power tempering and a local self-training interpretation. Stronger perturbations may require higher-order posterior information, while strongly multimodal posteriors require corrections beyond covariance. Future work includes scalable estimation of the posterior geometry in high-dimensional diffusion models, extensions to more general distributional perturbations, and understanding how the local training-space prediction evolves under finite optimization.

\bibliographystyle{apalike}
\bibliography{main}

\clearpage
\appendix

\section{Regularity, Notation, and Standard Posterior Identities}

\label{app:regularity}

We collect the assumptions needed to justify differentiation under the integral and interchanges of spatial, time, and temperature derivatives. For the exact identities, the relevant posterior moments and differentiated expectations are assumed finite. For power tempering near $\gamma=1$, a convenient sufficient condition is~\eqref{eq:lownoise_integrability}. It ensures that the normalized power family exists in a neighbourhood of one and that $\int p_0|\log p_0|^m<\infty$ for every fixed nonnegative integer $m$; the same logarithmic moments are integrable uniformly for powers in a smaller neighbourhood of one.

For Theorem~\ref{thm:lownoise}, we additionally require the stated $C^6$ regularity and positive lower bound on an open neighbourhood of the compact set $K$. The bounds are uniform on $K$ at fixed dimension. For the posterior-defined low-noise quantities alone, the global condition can be weakened to $\int p_0|\log p_0|<\infty$; we use~\eqref{eq:lownoise_integrability} to retain the normalized-power interpretation throughout. Appendix~\ref{app:lownoise} localizes the convolution integrals and derives the posterior-negentropy rate from these assumptions. No separate bound on $\nabla_x\nu_t$ is imposed.

We write $O_{C^k}(\sigma_t^m)$ when the remainder and all spatial derivatives through order $k$ are uniformly $O(\sigma_t^m)$ on the specified compact set. This derivative control is part of the expansion and is not inferred by differentiating a scalar big-$O$ bound.

For completeness, differentiating the Gaussian convolution gives the first two Tweedie identities~\eqref{eq:tweedie1}--\eqref{eq:tweedie2}; see, e.g., \citet{efron2011tweedie,meng2021higher}. In particular $B_t\succeq0$ for every exact score field because it is a rescaled posterior covariance.

\section{Proof of the Posterior-KL Characterization}

\label{app:klproof}

\begin{proof}[Proof of Theorem~\ref{thm:kl}]

For the power tilt, Proposition~\ref{prop:tilt} with $f=L=\log p_0$ gives
\begin{equation*}
h_t=\nabla_x\E[L(X_0)\mid X_t=x].
\end{equation*}
The normalized channel kernel, viewed as a density in $x_0$, is
\begin{equation*}
\Lambda^{\flat}_{t,x}(x_0)
=a_t^dK_t(x\mid x_0)
=\mathcal N\!\left(x_0;\frac{x}{a_t},\frac{\sigma_t^2}{a_t^2}I\right).
\end{equation*}
Bayes' rule gives
\begin{equation*}
\frac{\pi_{t,x}(x_0)}{\Lambda^{\flat}_{t,x}(x_0)}
=\frac{p_0(x_0)}{a_t^dp_t(x)}.
\end{equation*}
Taking the posterior expectation of the logarithm therefore yields
\begin{equation*}
G_t(x)=\E[L(X_0)\mid X_t=x]-\log p_t(x)-d\log a_t.
\end{equation*}
The last term is independent of $x$. Differentiating gives
\begin{equation*}
\nabla_xG_t(x)=h_t(x)-s_t(x),
\end{equation*}
which is~\eqref{eq:klmain}.
\end{proof}

The identity also clarifies why $G_t$ is a useful potential rather than merely an algebraic residual. It is nonnegative pointwise, yet only its spatial variation affects the score. Hence a posterior may be far from the flat-prior Gaussian while contributing no tempering defect if that mismatch is spatially constant, as in the Gaussian specialization.

\section{Proof of the Covariance Decomposition and Spectral Bounds}

\label{app:decompproof}

\begin{proof}[Proof of Theorem~\ref{thm:decomp}]

Let $m_t=\E[X_0\mid X_t=x]$ and $C_t=\Cov(X_0\mid X_t=x)$, and let $\Gamma_{t,x}=\mathcal N(m_t,C_t)$. Since $\Gamma_{t,x}$ matches the first two moments of $\pi_{t,x}$, the Gaussian cross-entropy term has the same expectation under $\pi_{t,x}$ and $\Gamma_{t,x}$. Consequently,
\begin{equation*}
\KL(\pi_{t,x}\|\Lambda^{\flat}_{t,x})
=\KL(\pi_{t,x}\|\Gamma_{t,x})
+\KL(\Gamma_{t,x}\|\Lambda^{\flat}_{t,x}).
\end{equation*}
The Gaussian KL is
\begin{align*}
\KL(\Gamma_{t,x}\|\Lambda^{\flat}_{t,x})
=\frac12\bigg[
&\frac{a_t^2}{\sigma_t^2}\|m_t-x/a_t\|^2
+\frac{a_t^2}{\sigma_t^2}\tr C_t-d\\
&-\log\det\!\left(\frac{a_t^2}{\sigma_t^2}C_t\right)
\bigg].
\end{align*}
Using the Tweedie identities~\eqref{eq:tweedie1}--\eqref{eq:tweedie2},
\begin{equation*}
\frac{a_t^2}{\sigma_t^2}\|m_t-x/a_t\|^2
=\sigma_t^2\|s_t(x)\|^2,
\qquad
\frac{a_t^2}{\sigma_t^2}C_t=B_t(x),
\end{equation*}
which gives~\eqref{eq:Gdecomp} with the definitions~\eqref{eq:phidef}--\eqref{eq:nudef}.

Differentiate~\eqref{eq:Gdecomp} and use Theorem~\ref{thm:kl}. Since $s_t=\nabla\log p_t$, its Jacobian is symmetric, and therefore
\begin{equation*}
\nabla_x\!\left(\frac{\sigma_t^2}{2}\|s_t\|^2\right)
=\sigma_t^2(\nabla_xs_t)s_t=(B_t-I)s_t.
\end{equation*}
Hence
\begin{equation*}
h_t-s_t=(B_t-I)s_t+\nabla_x(\Phi_t+\nu_t),
\end{equation*}
which rearranges to~\eqref{eq:hdecomp}.

\end{proof}

\paragraph{Spectral sandwich for strongly log-concave targets.}

This argument is a direct posterior covariance bound; related log-Hessian estimates along heat flow are developed by \citet{brigati2025heat}. Suppose $p_0\propto e^{-U}$ with $\alpha I\preceq\nabla^2U\preceq LI$. The posterior potential is
\begin{equation*}
V_{t,x}(x_0)=U(x_0)+\frac{\|x-a_tx_0\|^2}{2\sigma_t^2}+\mathrm{const},
\end{equation*}
so
\begin{equation*}
\left(\alpha+\frac{a_t^2}{\sigma_t^2}\right)I
\preceq \nabla^2V_{t,x}
\preceq
\left(L+\frac{a_t^2}{\sigma_t^2}\right)I.
\end{equation*}
The Brascamp--Lieb variance inequality \citep{brascamp1976extensions} applied to linear functionals gives the upper covariance bound
\begin{equation*}
C_t\preceq
\frac{1}{\alpha+a_t^2/\sigma_t^2}I.
\end{equation*}
For the lower bound, the Cram\'er--Rao information inequality gives $C_t\succeq J_t^{-1}$, where the posterior Fisher information satisfies
\begin{equation*}
J_t=\E_{\pi_{t,x}}[\nabla^2V_{t,x}(X_0)]
\preceq
\left(L+\frac{a_t^2}{\sigma_t^2}\right)I.
\end{equation*}
Thus
\begin{equation*}
\frac{1}{L+a_t^2/\sigma_t^2}I
\preceq C_t
\preceq
\frac{1}{\alpha+a_t^2/\sigma_t^2}I.
\end{equation*}
Multiplying by $a_t^2/\sigma_t^2$ proves~\eqref{eq:logconcave_bound}.

\section{Why Covariance Is Leading-Order at Low Noise}
\label{app:lownoise}

We prove Theorem~\ref{thm:lownoise} through an exact pointwise evolution equation for posterior negentropy. The heat-flow score and log-Hessian identities used to obtain it are standard; see, for example, \citet{brigati2025heat} for modern log-Hessian estimates along heat flow and their connection to score-based diffusion. All spatial remainders below are uniform on a fixed compact set, and $O_{C^k}$ controls spatial derivatives through order $k$. The argument also identifies the leading coefficients of the two score errors.

\begin{proof}[Proof of Theorem~\ref{thm:lownoise}]
\textbf{Gaussian-convolution coordinates.}
Suppress $t$ and put $y=x/a$ and $\tau=\sigma^2/a^2$. Let
\begin{equation*}
P_\tau f(y)=\E[f(y+\sqrt\tau Z)],\qquad
q_\tau=P_\tau p_0,\qquad L_\tau=\log q_\tau,
\end{equation*}
so that $p_t(x)=a^{-d}q_\tau(y)$. Define
\begin{equation*}
A_\tau=\frac{P_\tau(p_0\ell)}{q_\tau},\qquad
G_\tau=A_\tau-L_\tau,\qquad
g_\tau=\nabla_yL_\tau,\qquad H_\tau=\nabla_y^2L_\tau.
\end{equation*}
In these coordinates the normalized posterior covariance is $B_\tau=I+\tau H_\tau$. The exact decomposition~\eqref{eq:Gdecomp} reads
\begin{equation}
\nu_\tau=G_\tau-\frac\tau2\|g_\tau\|^2-\Phi_\tau,
\qquad
\Phi_\tau=\frac12\bigl[\tau\tr H_\tau-\log\det(I+\tau H_\tau)\bigr].
\label{eq:ln_negentropy_identity}
\end{equation}
Here $\nu_t(x)=\nu_\tau(y)$, $\Phi_t(x)=\Phi_\tau(y)$, and $\nabla_x=a^{-1}\nabla_y$.

\textbf{Localization and integrability.}
Choose compact sets $K\subset\operatorname{int}K_1\subset\operatorname{int}K_2\subset U$. For sufficiently small noise, $x/a_t\in K_1$ whenever $x\in K$. Choose a smooth cutoff $\chi$ supported in a compact subset of $U$ and equal to one near $K_2$. Set
\begin{equation*}
\widetilde\ell(u)=\chi(u)\ell(u)-(1-\chi(u))\sqrt{1+\|u\|^2},
\qquad w(u)=e^{\widetilde\ell(u)}.
\end{equation*}
The product $\chi\ell$ is extended by zero outside its support. Then $w=p_0$ near $K_2$, $w$ has exponential tails, and derivatives of $\widetilde\ell$ of orders one through six are globally bounded. Normalization of $w$ is unnecessary: multiplying a prior weight by a positive constant leaves its posterior, $G_\tau$, $\Phi_\tau$, and $\nu_\tau$ unchanged.

The power-integrability assumption implies $p_0|\ell|\in L^1$. Hence both $p_0-w$ and $p_0\ell-w\widetilde\ell$ are integrable and vanish on a neighbourhood of $K_2$. For either difference $f$ and every fixed multi-index $\alpha$, differentiation of the Gaussian kernel gives
\begin{equation*}
\sup_{y\in K_1}|\partial_y^\alpha P_\tau f(y)|
\le C_\alpha\tau^{-N_\alpha}e^{-c/\tau}\|f\|_{L^1}.
\end{equation*}
The positive distance to the support of $f$ supplies the exponential factor; the polynomial factors from kernel differentiation are absorbed into a weaker Gaussian exponential. Both convolved weights are bounded below on $K_1$ for small $\tau$. Equation~\eqref{eq:ln_negentropy_identity} therefore shows that the original and extended $\nu_\tau$ differ by $O_{C^1}(e^{-c'/\tau})$: only three spatial derivatives of $q_\tau$ and one of $P_\tau(p_0\ell)$ are needed. The score fields have the same localization property. It is thus sufficient to prove the expansions for $w$ and then transfer them to $p_0$ on $K_1$. In the remainder of the proof, write $\ell$ for the extended log-weight and $q_\tau=P_\tau w$.

\textbf{Uniform spatial regularity.}
Bounded derivatives of $\ell$ imply $|\partial^\alpha w|\le C_\alpha w$ for $|\alpha|\le6$. Commuting spatial derivatives with convolution yields
\begin{equation}
|\partial^\alpha q_\tau|\le C_\alpha q_\tau,\qquad |\alpha|\le6.
\label{eq:ln_derivative_bounds}
\end{equation}
Consequently all spatial derivatives of $L_\tau$ of orders one through six are uniformly bounded. In particular $B_\tau=I+O(\tau)$ has a uniformly bounded inverse for small $\tau$. We also have
\begin{equation}
L_\tau-\ell=O_{C^4}(\tau).
\label{eq:ln_c4_convergence}
\end{equation}
To verify the derivative count, for $|\alpha|\le4$ use
\begin{equation*}
\partial^\alpha(q_\tau-w)=\frac12\int_0^\tau P_r(\Delta\partial^\alpha w)\,\dd r.
\end{equation*}
The integrand is bounded by a constant times $q_r$, and the globally Lipschitz log-weight makes $q_r/w$ uniformly bounded above and below for small $r$. Division by $w$ and the logarithmic chain rule prove~\eqref{eq:ln_c4_convergence}, using at most six derivatives of the initial log-weight.

\textbf{Pointwise evolution of posterior negentropy.}
Introduce
\begin{equation*}
\mathcal D_\tau=\partial_\tau-\tfrac12\Delta_y-g_\tau\cdot\nabla_y.
\end{equation*}
Since $q_\tau$ and $P_\tau(w\ell)$ solve the heat equation, the quotient rule and spatial differentiation give
\begin{align}
\partial_\tau L_\tau&=\tfrac12\Delta_yL_\tau+\tfrac12\|g_\tau\|^2,\nonumber\\
\mathcal D_\tau A_\tau&=0,
&\mathcal D_\tau G_\tau&=\tfrac12\|g_\tau\|^2,\nonumber\\
\mathcal D_\tau g_\tau&=0,
&\mathcal D_\tau H_\tau&=H_\tau^2.
\label{eq:ln_heat_evolution}
\end{align}
The product rule for $\mathcal D_\tau$ includes $-\nabla f\cdot\nabla h$. Therefore
\begin{equation*}
\mathcal D_\tau\!\left(\frac\tau2\|g_\tau\|^2\right)
=\frac12\|g_\tau\|^2-\frac\tau2\|H_\tau\|_{\mathrm F}^2.
\end{equation*}
Define
\begin{align}
Q_\tau
&=\sum_{k=1}^d\tr\!\left[B_\tau^{-1}(\partial_kH_\tau)B_\tau^{-1}(\partial_kH_\tau)\right]\nonumber\\
&=\sum_{k=1}^d\left\|B_\tau^{-1/2}(\partial_kH_\tau)B_\tau^{-1/2}\right\|_{\mathrm F}^2\ge0.
\label{eq:ln_source}
\end{align}
From~\eqref{eq:ln_heat_evolution}, $\mathcal D_\tau B_\tau=H_\tau+\tau H_\tau^2=H_\tau B_\tau$. The matrix chain rule gives
\begin{align*}
\mathcal D_\tau\log\det B_\tau
&=\tr(B_\tau^{-1}\mathcal D_\tau B_\tau)
+\frac12\sum_k\tr\!\left[B_\tau^{-1}(\partial_kB_\tau)B_\tau^{-1}(\partial_kB_\tau)\right]\\
&=\tr H_\tau+\frac{\tau^2}{2}Q_\tau.
\end{align*}
This calculation does not assume that $B_\tau$ and $\partial_kH_\tau$ commute. It follows that
\begin{equation*}
\mathcal D_\tau\Phi_\tau=\frac\tau2\|H_\tau\|_{\mathrm F}^2-\frac{\tau^2}{4}Q_\tau.
\end{equation*}
Substituting into~\eqref{eq:ln_negentropy_identity} cancels the score and Hessian terms exactly:
\begin{equation}
\mathcal D_\tau\nu_\tau=\frac{\tau^2}{4}Q_\tau,
\qquad \nu_0=0.
\label{eq:ln_negentropy_pde}
\end{equation}
The initial value is the continuous limit of~\eqref{eq:ln_negentropy_identity}; no rate of posterior Gaussianity has been used. This is the structural identity used in the proof. In contrast with classical de Bruijn/entropy-production formulas, which concern quantities integrated over the evolving density \citep{stam1959information,costa1985entropy}, equation~\eqref{eq:ln_negentropy_pde} is pointwise in the noisy coordinate $y$ and concerns the moment-matched-Gaussian divergence of the denoising posterior. We are not aware of this exact posterior-negentropy evolution identity being stated in this form in the prior literature; the constituent heat-flow identities themselves are standard.

\textbf{The expansion and its spatial derivative.}
Equations~\eqref{eq:ln_c4_convergence} and~\eqref{eq:ln_source} imply
\begin{equation*}
Q_r=Q_0+O_{C^1}(r),\qquad Q_0=\|\nabla^3\ell\|_{\mathrm F}^2.
\end{equation*}
The tensor norm sums over all ordered indices, and $Q_0$ has three bounded spatial derivatives. Multiplying~\eqref{eq:ln_negentropy_pde} by $q_\tau$ gives
\begin{equation*}
(\partial_\tau-\tfrac12\Delta_y)(q_\tau\nu_\tau)
=\frac{\tau^2}{4}q_\tau Q_\tau.
\end{equation*}
The exponential tails of the extended weight justify Duhamel's formula with zero initial data:
\begin{equation}
\nu_\tau(y)=\frac{1}{4q_\tau(y)}\int_0^\tau r^2P_{\tau-r}(q_rQ_r)(y)\,\dd r.
\label{eq:ln_duhamel}
\end{equation}
For explicit control of differentiated remainders, define
\begin{equation*}
T_{\tau,r}f=\frac{P_{\tau-r}(q_rf)}{q_\tau}.
\end{equation*}
Since $P_{\tau-r}q_r=q_\tau$, this is a positive averaging operator. Its derivative satisfies
\begin{equation*}
\partial_iT_{\tau,r}f
=T_{\tau,r}(\partial_if+g_{r,i}f)-g_{\tau,i}T_{\tau,r}f.
\end{equation*}
Boundedness of $g_r$ makes $T_{\tau,r}$ uniformly bounded on $C^1$. Moreover,
\begin{equation}
T_{\tau,r}Q_0=Q_0+O_{C^1}(\tau-r).
\label{eq:ln_averaging_expansion}
\end{equation}
Indeed, for $s=\tau-r$, $f=q_r$, and $b=Q_0$, the heat-semigroup commutator identity is
\begin{equation*}
P_s(fb)-bP_sf
=\frac12\int_0^sP_{s-v}\!\left[(\Delta b)P_vf+2\nabla b\cdot\nabla P_vf\right]\,\dd v.
\end{equation*}
Divide by $q_\tau$ and differentiate once. Derivatives of $b$ through order three are bounded, while derivatives of $P_vf=q_{r+v}$ divided by itself are bounded by~\eqref{eq:ln_derivative_bounds}. Both the value and the first derivative are therefore $O(s)$, proving~\eqref{eq:ln_averaging_expansion}. Thus
\begin{equation*}
T_{\tau,r}Q_r=Q_0+O_{C^1}(\tau),\qquad 0\le r\le\tau.
\end{equation*}
Integrating in~\eqref{eq:ln_duhamel} yields
\begin{equation*}
\nu_\tau(y)=\frac{\tau^3}{12}\|\nabla^3\ell(y)\|_{\mathrm F}^2+O_{C^1}(\tau^4).
\end{equation*}
The coefficient in the value expansion is consistent with the classical Edgeworth/cumulant formula for negentropy \citep{jones1987projection,comon1994ica}. Indeed, after centering and whitening the small-$\tau$ posterior, its standardized third-cumulant tensor is $\tau^{3/2}\nabla^3\ell+O(\tau^{5/2})$, so the classical quadratic third-cumulant contribution gives $\tau^3\|\nabla^3\ell\|_{\mathrm F}^2/12$; fourth-cumulant contributions enter at the next order. The pointwise PDE derivation above additionally provides the compact-uniform $C^1$ remainder required here. In particular $\nabla_y\nu_\tau=O(\tau^3)$. Localization transfers this expansion, with its derivative, to the original density on $K_1$.

\textbf{Covariance residual and the original coordinates.}
The matrix logarithm expansion and~\eqref{eq:ln_c4_convergence} give
\begin{equation*}
\Phi_\tau=\frac{\tau^2}{4}\tr(H_\tau^2)+O_{C^1}(\tau^3)
=\frac{\tau^2}{4}\|\nabla^2\ell\|_{\mathrm F}^2+O_{C^1}(\tau^3).
\end{equation*}
The derivative remainder can also be checked directly from
$\partial_i\Phi_\tau=\frac12\tr[(I-B_\tau^{-1})\partial_iB_\tau]$.
Using $h_t-B_ts_t=a_t^{-1}\nabla_y(\Phi_\tau+\nu_\tau)$ and $(B_t-I)s_t=\sigma_t^2(\nabla_xs_t)s_t$ gives
\begin{align*}
\nu_t(x)
&=\frac{\sigma_t^6}{12a_t^6}\|\nabla^3\ell(x/a_t)\|_{\mathrm F}^2+O_{C^1_x}(\sigma_t^8),\\
\nabla_x\nu_t(x)
&=\frac{\sigma_t^6}{12a_t^7}\nabla\|\nabla^3\ell\|_{\mathrm F}^2(x/a_t)+O(\sigma_t^8),\\
h_t(x)-B_t(x)s_t(x)
&=\frac{\sigma_t^4}{4a_t^5}\nabla\|\nabla^2\ell\|_{\mathrm F}^2(x/a_t)+O(\sigma_t^6),\\
h_t(x)-s_t(x)
&=\frac{\sigma_t^2}{a_t^3}\nabla^2\ell(x/a_t)\nabla\ell(x/a_t)+O(\sigma_t^4).
\end{align*}
All remainders are uniform for $x\in K$, with $a_t$ bounded above and away from zero. Finally, $a_t=1+O(\sigma_t^2)$ allows replacement of $x/a_t$ by $x$ and of the prefactors $a_t^{-j}$ by one at the stated orders. This proves~\eqref{eq:lownoise} and~\eqref{eq:lownoise_negentropy}. More precisely,
\begin{align}
h_t-B_ts_t
&=\frac{\sigma_t^4}{4}\nabla\|\nabla^2\ell\|_{\mathrm F}^2+O(\sigma_t^6),\nonumber\\
h_t-s_t
&=\sigma_t^2\nabla^2\ell\,\nabla\ell+O(\sigma_t^4).
\label{eq:ln_leading_score_errors}
\end{align}
\end{proof}

Equation~\eqref{eq:ln_leading_score_errors} shows that the fourth-order covariance residual is led by the spatial covariance-mismatch term $\nabla\Phi_t$, while $\nabla\nu_t$ enters only at sixth order. The displayed coefficients can vanish, so these compact-uniform expansions do not assert matching lower bounds at every point. For a Gaussian prior, $\nabla^3\ell=0$ and $\nabla^2\ell$ is spatially constant, consistently with exact covariance propagation. Smooth multimodal densities are also covered locally as noise tends to zero; singular targets, global uniformity over all of $\R^d$, and intermediate-noise posterior ambiguity are not consequences of this theorem.

\section{Beyond Infinitesimal Temperature}

\label{app:finite}

This appendix concerns the distinct asymptotic axis $\delta=\gamma-1\to0$. Proposition~\ref{prop:finite} is proved in the main text; here we record the error decomposition and the Gaussian specialization.

From~\eqref{eq:cumulant} and the covariance approximation~\eqref{eq:covsampler},
\begin{align*}
\nabla_x\log q_{\gamma,t}-s_t^{\rm cov}(\gamma)
&=\delta(h_t-B_ts_t)
+\frac{\delta^2}{2}\nabla_x\Var(L(X_0)\mid X_t)\\
&\quad+O_t(\delta^3).
\end{align*}
Theorem~\ref{thm:lownoise} makes the first term $O(|\delta|\sigma_t^4)$; its posterior-negentropy component is $O(|\delta|\sigma_t^6)$. For the second term, apply the localized Gaussian-convolution expansion to $p_0$, $p_0L$, and $p_0L^2$. The power-integrability condition controls their tails, and the local smoothness controls the expansion in $C^1$. In the coordinates $y=x/a_t$, $\tau=\sigma_t^2/a_t^2$, this gives
\begin{equation*}
\Var(L(X_0)\mid X_t=x)
=\frac{\sigma_t^2}{a_t^2}\|\nabla L(x/a_t)\|^2+O_{C^1_x}(\sigma_t^4).
\end{equation*}
Its spatial gradient is therefore $O(\sigma_t^2)$. Combining these estimates with the conditional-cumulant remainder proves~\eqref{eq:finiteerror}.

For $p_0=\mathcal N(0,\Sigma)$, clean powering gives covariance $\Sigma/\gamma$. After forward noising,
\begin{equation*}
q_{\gamma,t}=\mathcal N\!\left(0,\frac{a_t^2}{\gamma}\Sigma+\sigma_t^2I\right),
\end{equation*}
so
\begin{equation}
\nabla_x\log q_{\gamma,t}(x)
=-\gamma(a_t^2\Sigma+\gamma\sigma_t^2I)^{-1}x.
\label{eq:gaussian_finite_score_app}
\end{equation}
Under a variance-preserving parameterization, choose $t'$ by~\eqref{eq:shift_time}. Then
\begin{equation*}
a_{t'}^2\Sigma+\sigma_{t'}^2I
=\frac{a_t^2\Sigma+\gamma\sigma_t^2I}{a_t^2+\gamma\sigma_t^2}.
\end{equation*}
Substituting this covariance into the original Gaussian score gives
\begin{equation*}
\frac{\gamma}{a_t^2+\gamma\sigma_t^2}s_{t'}(x)
=-\gamma(a_t^2\Sigma+\gamma\sigma_t^2I)^{-1}x,
\end{equation*}
which proves~\eqref{eq:shiftrescale}. Differentiating~\eqref{eq:gaussian_finite_score_app} at $\gamma=1$ recovers $B_ts_t$.

\paragraph{Gaussian recovery of temporal rescaling.}

For $\Sigma=\tau^2I$, $B_t=r_tI$ with
\begin{equation*}
r_t=\frac{a_t^2\tau^2}{a_t^2\tau^2+\sigma_t^2}.
\end{equation*}
This is the infinitesimal resolved-variance factor obtained by differentiating the isotropic Gaussian temporal rescaling used in TSR at unit temperature. In the present framework it appears as the isotropic specialization of the posterior-covariance response. TSR's local-temperature construction and the global target in~\eqref{eq:intro_clean_power} agree on a single Gaussian but have different mixture semantics in general.

\section{Frozen-Covariance Objective Details}

\label{app:regularization}

The covariance potential introduced in Section~\ref{sec:synthetic} is defined using the detached reference-model marginals and posterior geometry frozen at the reference checkpoint $\theta_r$:
\begin{equation}
V_{\rm cov}(\theta)
=
\frac12
\int
\rho(t)
\E_{p_{\theta_r,t}}
\left[
s_\theta(X,t)^\top
B_t^{\theta_r}(X)
s_\theta(X,t)
\right]
\dd t.
\label{eq:app_vcov}
\end{equation}
Only $s_\theta$ is differentiated; both the sampling measure $p_{\theta_r,t}$ and the covariance field $B_t^{\theta_r}$ are held fixed. Its gradient at the reference checkpoint is therefore
\begin{equation*}
\nabla_\theta V_{\rm cov}(\theta_r)
=
\int
\rho(t)
\E_{p_{\theta_r,t}}
\left[
(\partial_\theta s_{\theta_r})^\top
B_t^{\theta_r}
s_{\theta_r}
\right]
\dd t,
\end{equation*}
which is exactly the posterior-covariance component of the local synthetic-training pullback. For a Gaussian reference distribution the posterior residual vanishes, and hence this covariance term gives the complete first-order frozen potential.

\paragraph{Time weighting and $\epsilon$-parameterization.}
To make the dependence on timestep sampling explicit, let $t\sim p(t)$ and write
\begin{equation*}
\rho(t)=p(t)\lambda_s(t),
\end{equation*}
where $\lambda_s(t)$ denotes any additional weighting of the score-space objective. For an $\epsilon$-prediction model,
\begin{equation*}
s_\theta(x,t)
=
-\frac{\epsilon_\theta(x,t)}{\sigma_t}.
\end{equation*}
Substituting this relation into~\eqref{eq:app_vcov} gives
\begin{equation}
V_{\rm cov}(\theta)
=
\frac12
\E_{t\sim p(t)}
\E_{p_{\theta_r,t}}
\left[
\frac{\lambda_s(t)}{\sigma_t^2}
\epsilon_\theta(X,t)^\top
B_t^{\theta_r}(X)
\epsilon_\theta(X,t)
\right].
\label{eq:app_vcov_epsilon}
\end{equation}
Thus, when the objective is specified with a fixed score-space weighting $\lambda_s(t)$, conversion to $\epsilon$-parameterization introduces the factor $1/\sigma_t^2$. In particular, a time-unweighted score-space objective, $\lambda_s(t)=1$, corresponds to an $\epsilon$-space weighting proportional to
\begin{equation*}
\frac{p(t)}{\sigma_t^2}.
\end{equation*}
This conversion should be distinguished from the weighting implicit in standard $\epsilon$-prediction DSM. Since
\begin{equation*}
\|\epsilon_\theta-\epsilon^\star\|^2
=
\sigma_t^2
\|s_\theta-s^\star\|^2,
\end{equation*}
an unweighted $\epsilon$-prediction loss corresponds to the score-space weighting $\lambda_s(t)=\sigma_t^2$. Under this training-matched choice, the $1/\sigma_t^2$ factor in~\eqref{eq:app_vcov_epsilon} cancels, giving
\begin{equation*}
V_{\rm cov}^{(\epsilon)}(\theta)
=
\frac12
\E_{t\sim p(t)}
\E_{p_{\theta_r,t}}
\left[
\epsilon_\theta(X,t)^\top
B_t^{\theta_r}(X)
\epsilon_\theta(X,t)
\right].
\end{equation*}
The appropriate implementation therefore depends on the timestep distribution and weighting convention of the underlying DSM objective.

\paragraph{Isotropic approximation.}

If the posterior-covariance operator is further approximated by the identity,

$B_t^{\theta_r}(x)\approx I$, the covariance potential reduces to
\begin{equation*}
V_{\rm iso}(\theta)
=
\frac12
\int
\rho(t)
\E_{p_{\theta_r,t}}
\left[
\|s_\theta(X,t)\|^2
\right]
\dd t,
\end{equation*}
which is ordinary score-norm regularization. In $\epsilon$-parameterization, the corresponding expression is
\begin{equation*}
V_{\rm iso}(\theta)
=
\frac12
\E_{t\sim p(t)}
\left[
\frac{\lambda_s(t)}{\sigma_t^2}
\E_{p_{\theta_r,t}}
\|\epsilon_\theta(X,t)\|^2
\right],
\end{equation*}
with the same cancellation of $\sigma_t^{-2}$ when $\lambda_s(t)=\sigma_t^2$.

\section{Temporal-Consistency Identity}

\label{app:fp}

A separate line of work studies whether learned score fields form a temporally coherent diffusion family; FP-Diffusion derives and regularizes the corresponding score Fokker--Planck residual \citep{lai2023fp}. Here the connection is only an interpretation of one component of the clean-tempering error.

The full infinitesimal failure of naive noisy-score tempering is
\begin{equation*}
h_t-s_t=(B_t-I)s_t+r_t.
\end{equation*}
For an exact score, $\nabla_xs_t$ is symmetric, and hence
\begin{equation*}
(B_t-I)s_t=\sigma_t^2(\nabla_xs_t)s_t=\frac{\sigma_t^2}{2}\nabla_x\|s_t\|^2.
\end{equation*}
For the variance-preserving SDE, write the score Fokker--Planck equation as
\begin{equation*}
\partial_t s_t=\nabla_x\mathcal L_t[s_t], \qquad
\mathcal L_t[s]=\frac{\beta(t)}{2}\left(\nabla\!\cdot s+\|s\|^2+x^\top s+d\right).
\end{equation*}
Let $\widetilde s_t^\delta=(1+\delta)s_t$ and let $\mathfrak R_{\delta,t}$ denote the score-Fokker--Planck residual of this naively rescaled family. The time derivative, divergence, and linear drift terms scale linearly and cancel against the exact equation at first order, whereas the quadratic term contributes one additional copy. Therefore
\begin{align}
\left.\partial_\delta\mathfrak R_{\delta,t}\right|_{\delta=0}
&=-\frac{\beta(t)}{2}\nabla_x\|s_t\|^2,\nonumber\\
(B_t-I)s_t
&=-\frac{\sigma_t^2}{\beta(t)}\left.\partial_\delta\mathfrak R_{\delta,t}\right|_{\delta=0}.
\label{eq:FPidentity}
\end{align}
Thus the first-order temporal inconsistency of naive scalar score rescaling identifies exactly the covariance component $(B_t-I)s_t$, not the full noncommutation error. For Gaussians $r_t=0$, while under Theorem~\ref{thm:lownoise} the residual is higher order at low noise. The additional posterior residual at intermediate noise remains outside this identity.

\section{Proof of the Synthetic DSM Tangent}

\label{app:synthetic}

\begin{proof}[Proof of Theorem~\ref{thm:syn}]

Write $q_{\delta,t}:=q_{1+\delta,t}$ and
\begin{equation*}
s_{q,\delta}(x,t):=\nabla_x\log q_{\delta,t}(x).
\end{equation*}
The synthetic population DSM objective is
\begin{equation*}
R_{\rm syn}(\theta)
=\frac12\int\rho(t)\E_{q_{\delta,t}}
\left[\|s_\theta(X,t)-s_{q,\delta}(X,t)\|^2\right]\dd t.
\end{equation*}
The synthetic distribution is generated from the fixed reference checkpoint $\theta_r$. Hence, when differentiating with respect to $\theta$, both $q_{\delta,t}$ and its target score $s_{q,\delta}$ are held fixed. Differentiation under the expectation gives
\begin{equation}
\nabla_\theta R_{\rm syn}(\theta)
=\int\rho(t)\E_{q_{\delta,t}}
\left[(\partial_\theta s_\theta)^\top(s_\theta-s_{q,\delta})\right]\dd t.
\label{eq:app_Rsyn_grad}
\end{equation}
At the self-consistent reference checkpoint, $s_{\theta_r}=s_t$. Proposition~\ref{prop:finite} gives the local score expansion
\begin{equation}
s_{q,\delta}(x,t)=s_t(x)+\delta h_t(x)+O_t(\delta^2).
\label{eq:app_tempered_score_expansion}
\end{equation}
Substituting~\eqref{eq:app_tempered_score_expansion} into~\eqref{eq:app_Rsyn_grad} at $\theta=\theta_r$ yields
\begin{equation}
\nabla_\theta R_{\rm syn}(\theta_r)
=-\delta\int\rho(t)\E_{q_{\delta,t}}
\left[(\partial_\theta s_{\theta_r})^\top h_t\right]\dd t
+O(\delta^2).
\label{eq:app_grad_qmeasure}
\end{equation}
Because $q_{\delta,t}$ is a smooth perturbation of $p_{\theta_r,t}$ around $\delta=0$, every integrable test field $F_t$ satisfying the regularity conditions of Appendix~\ref{app:regularity} obeys
\begin{equation*}
\E_{q_{\delta,t}}[F_t(X)]
=\E_{p_{\theta_r,t}}[F_t(X)]+O(\delta).
\end{equation*}
The expectation difference in~\eqref{eq:app_grad_qmeasure} is already multiplied by $\delta$, so replacing $q_{\delta,t}$ by $p_{\theta_r,t}$ changes the result only by $O(\delta^2)$. Therefore
\begin{equation*}
\nabla_\theta R_{\rm syn}(\theta_r)
=-\delta\int\rho(t)\E_{p_{\theta_r,t}}
\left[(\partial_\theta s_{\theta_r})^\top h_t\right]\dd t
+O(\delta^2),
\end{equation*}
which is~\eqref{eq:syngrad}.

For $\delta=0$, $q_{0,t}=p_{\theta_r,t}$ and $s_{q,0}=s_{\theta_r}$, so the integrand in~\eqref{eq:app_Rsyn_grad} vanishes pointwise. Hence exact population self-sampling produces zero systematic first-order gradient.

\end{proof}

\begin{remark}[Why the sampler is frozen]

The theorem describes the standard self-training procedure in which a synthetic dataset is generated from a reference checkpoint and then used for optimization. The sampling distribution is therefore held fixed while differentiating the training loss. Differentiating through a sampler that changes simultaneously with $\theta$ would produce additional terms and is a different optimization problem.

\end{remark}

Equations~\eqref{eq:gradVcov}--\eqref{eq:synthetic_potential} in the main text then follow by freezing the measure and the fields $B_t,r_t$ at $\theta_r$. This frozen construction is a semi-gradient designed to reproduce the local synthetic-training direction; it is not a claim that $V_{\rm full}$ is the full parameter derivative of a model-distribution entropy.

\section{Additional Experiments}

\label{app:extraexp}

\subsection{Reverse-SDE endpoint test}

On an $8$D anisotropic Gaussian with eigenvalues $0.05$--$20$, we integrate the reverse SDE using the approximate score fields and evaluate KL to the exact power-tempered endpoint. At $\gamma=1.3$, naive score scaling gives endpoint KL $9.25\times10^{-2}$, the covariance first-order field $1.04\times10^{-3}$, and inclusion of the second temperature cumulant $2.13\times10^{-5}$. This experiment shows that the local field hierarchy propagates to an endpoint distribution in the regime where the perturbation remains controlled; the result is shown in the right panel of Figure~\ref{fig:appstructural}.

\subsection{Numerical check of the temporal-consistency identity}

\begin{figure}[h]
\centering
\includegraphics[width=\linewidth]{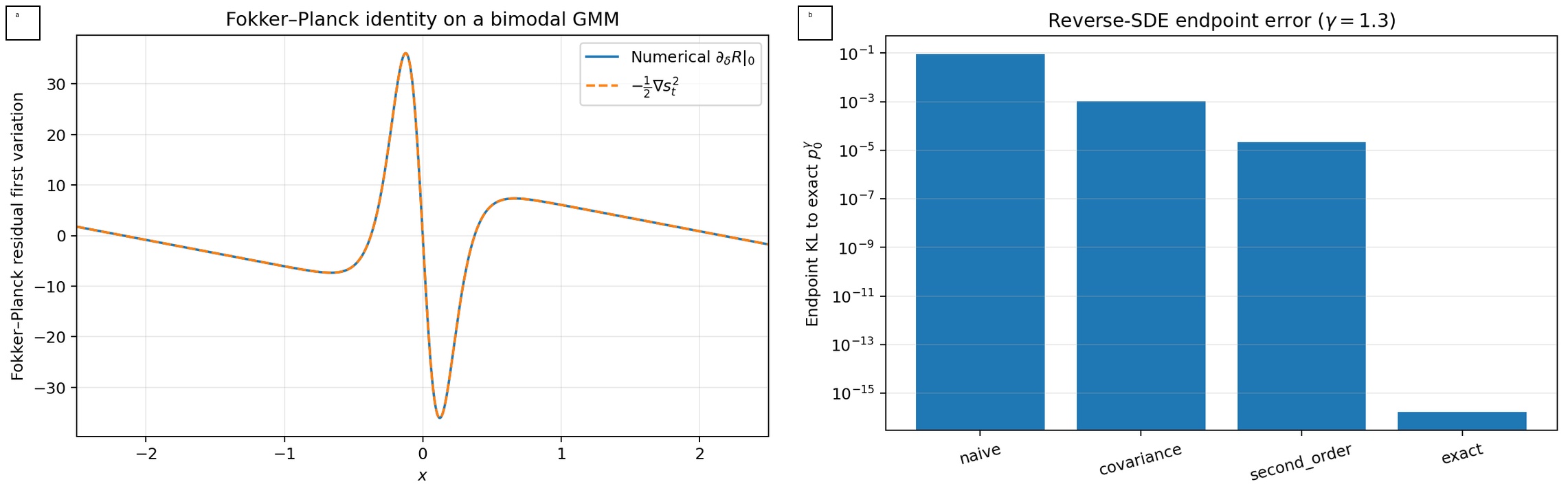}
\caption{Additional structural validation: Fokker--Planck first variation (left) and reverse-SDE endpoint KL (right).}
\label{fig:appstructural}
\end{figure}

On the bimodal GMM, we independently finite-difference the temporal score derivative and the nonlinear FP operator rather than substituting~\eqref{eq:FPidentity}. The first-variation field agrees with $-\frac12\nabla s_t^2$ with relative weighted RMS error $3.8\times10^{-5}$; the induced relation to $B_ts_t-s_t$ has relative error $1.6\times10^{-5}$. The left panel of Figure~\ref{fig:appstructural} shows this numerical identity check.

\begin{figure}[t]
\centering
\includegraphics[width=\linewidth]{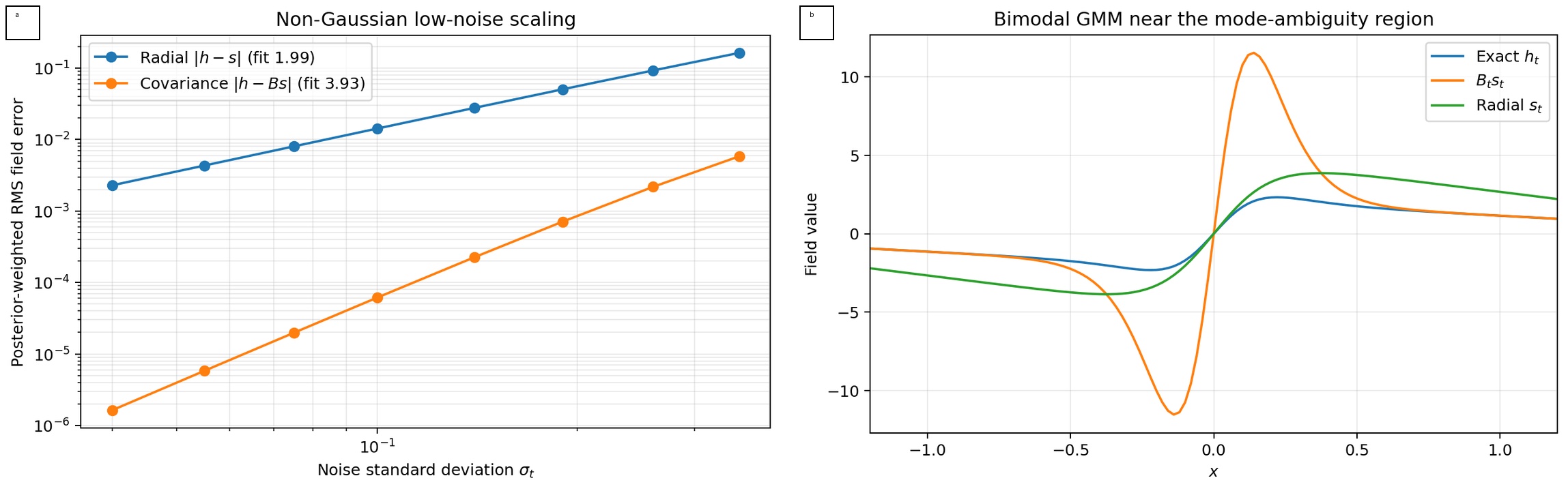}
\caption{\textbf{Covariance and defect regimes.} Left: for a smooth non-Gaussian density, naive radial error is second order in noise while the posterior-covariance residual is fourth order. Right: at intermediate noise in a bimodal mixture, unresolved mode identity produces a large posterior residual and covariance-only propagation can substantially overshoot the exact field.}
\label{fig:regimes}
\end{figure}

\begin{figure}[t]
\centering
\includegraphics[width=\linewidth]{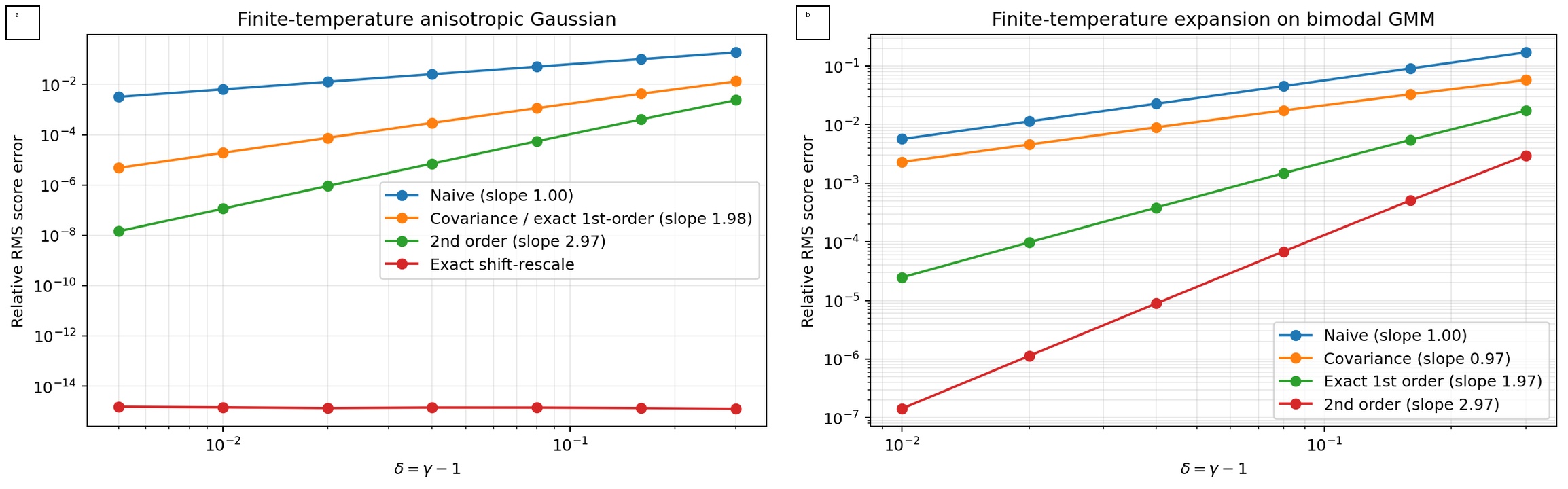}
\caption{\textbf{Finite-strength hierarchy.} Left: for anisotropic Gaussian, covariance is exact first-order tangent and second cumulant raises error order in $|\gamma-1|$; the Gaussian shift--rescale specialization is numerically exact. Right: on a GMM, covariance remains first order because posterior residual is nonzero, whereas exact tangent and second cumulant recover the predicted second- and third-order behavior.}
\label{fig:finite}
\end{figure}

\subsection{Additional learned-Gaussian checks}
The learned-Gaussian experiment in Section~\ref{sec:learnedapp} separates the finite-sampler self-training signal from the controlled score-scale distortion. With paired synthetic pools,
\[
\Delta g_{\rm syn}
=
g_{\rm syn}(c=1.1)-g_{\rm syn}(c=1),
\]
the covariance frozen-potential direction has cosine $0.9851\pm0.0071$ with the incremental synthetic gradient over nine seeds. On a more anisotropic rotated Gaussian with covariance eigenvalues $6$ and $0.3$, the covariance alignment remains $0.981\pm0.013$. After $250$ synthetic Adam steps in the main Gaussian setting, the full checkpoint displacement retains cosine $0.701\pm0.034$ with the initial covariance direction. These checks support the two localities in the application: the sampler distortion must remain close to the assumed power perturbation, and finite optimization can progressively move away from the tangent defined at $\theta_r$.

\subsection{MNIST-PCA: non-Gaussian supporting experiment}

\label{app:mnistpca}

We use the canonical MNIST split of $60{,}000$ training and $10{,}000$ test images, but train the diffusion model in a $100$-dimensional PCA representation fitted only on the training images. The retained coordinates explain $91.425\%$ of training-set pixel variance. The diffusion network is an $\epsilon$-prediction MLP with a $48$-dimensional sinusoidal time embedding and three $192$-unit SiLU hidden layers, trained with a cosine schedule over $100$ noise levels. Each reference model is trained for $9{,}500$ AdamW updates. Generated latent samples are inverse transformed to $28\times28$ images and evaluated against the untouched MNIST test set using Fr\'echet distance in the penultimate feature space of a separately trained LeNet classifier.

\paragraph{Controlled covariance-direction test.}

The theory-facing synthetic branch uses paired $20$-step samplers with common initial noise. The reference sampler uses score/noise multiplier $c=1$, while the controlled branch uses $c=1.05$. We isolate the incremental synthetic-training signal as
\[
\Delta g_{\rm syn}
=
g_{\rm syn}(c=1.05)-g_{\rm syn}(c=1).
\]
The comparison is made under the detached reference-model measure, matching Theorem~\ref{thm:syn}. Across five independently trained models, the covariance frozen-potential direction satisfies
\[
\cos\!\left(
\Delta g_{\rm syn},
-\nabla_\theta V_{\rm cov}
\right)
=
0.843\pm0.020.
\]
Thus a substantial part of the controlled synthetic-training perturbation is captured by the posterior-covariance pullback even on this learned, multimodal, non-Gaussian representation. Unlike the Gaussian setting, this agreement is not expected to be exact because the posterior residual need not vanish.

\paragraph{Endpoint negative extrapolation.}

Synthetic fine-tuning uses $2{,}000$ controlled samples and $200$ Adam updates. The reference model has LeNet feature FID $85.95\pm6.41$. Controlled synthetic fine-tuning worsens it to $113.56\pm5.75$, whereas extrapolating opposite to the resulting displacement with $w=1$ gives $74.24\pm6.17$. The paired change relative to each seed's base is $-11.71\pm1.52$ FID points, with improvement in all five runs. This endpoint result is consistent with the sign implied by the positively aligned controlled covariance direction, while the gradient-level cosine above is the more direct mechanistic test.

\paragraph{Frozen covariance objective.}

We separately evaluate the covariance objective~\eqref{eq:Jcov} using the detached reference-model measure for the covariance term. With the same fixed coefficient $\mu=0.3$ across five runs, the paired evaluation gives a base FID of $84.09\pm6.34$, an ordinary real-data continuation FID of $79.01\pm5.50$, and a frozen covariance-objective FID of $69.63\pm6.48$. Relative to the matched continuation-training checkpoint, the covariance objective changes FID by $-9.38\pm5.37$ points. We include this as a supporting endpoint check that the covariance potential can improve the reference model, while the central MNIST-PCA evidence remains the controlled synthetic-gradient alignment. The two endpoint experiments use separately paired evaluation streams and therefore have slightly different base FIDs; Table~\ref{tab:mnistpca_extra} summarizes the results.

\begin{table}[t]

\caption{\textbf{MNIST-PCA endpoint results.} Mean $\pm$ s.d.\ over five runs.}

\label{tab:mnistpca_extra}

\begin{center}
\begin{tabular}{lr}
\toprule
Method & LeNet FID $\downarrow$ \\
\midrule
\multicolumn{2}{l}{\textit{Synthetic-training branch}} \\
Base & $85.95\pm6.41$ \\
Synthetic FT & $113.56\pm5.75$ \\
Neg.\ extrapolation & $\mathbf{74.24\pm6.17}$ \\
\midrule
\multicolumn{2}{l}{\textit{Covariance-objective branch}} \\
Base & $84.09\pm6.34$ \\
Real-data continuation & $79.01\pm5.50$ \\
Covariance objective & $\mathbf{69.63\pm6.48}$ \\
\bottomrule
\end{tabular}
\end{center}

\end{table}

\section{Extended Related Work}
\label{app:related_work}

\paragraph{Temperature sampling and powered distributions.}

Temperature control is nontrivial in diffusion and flow models because scores or velocities are defined over noisy intermediate distributions rather than the clean law. Temporal Score Rescaling (TSR) \citep{xu2026tsr} shows, in the isotropic Gaussian setting, that a scalar clean temperature change induces time-dependent score rescaling. Our characterization recovers this factor and extends it to anisotropic Gaussians, where the response becomes matrix-valued, while separating covariance effects from spatial covariance mismatch and non-Gaussian posterior structure. For smooth positive densities, we further establish low-noise accuracy and the order separation of these corrections. Li et al.\ \citep{li2026diversify} study the global target $p_0^\gamma$, showing that naive score scaling can reweight modes while distorting within-mode variance, and propose variance-corrective time shifting. We instead characterize the noisy-score field obtained by diffusing the powered clean distribution. Related work also considers temperature changes through annealed sampling dynamics. Progressive Inference-Time Annealing (PITA) combines Boltzmann-temperature annealing with diffusion smoothing and Feynman--Kac/SMC corrections \citep{akhoundsadegh2025pita}. In contrast, we fix the forward diffusion channel and characterize the noisy-score field induced by globally powering the clean distribution. The Gaussian finite-temperature specialization recovers the corresponding shift--rescale behavior, while posterior residuals and conditional cumulants describe departures from this picture for non-Gaussian targets.

\paragraph{Guidance and intermediate-distribution mismatch.}

Similar issues arise when endpoint density transformations are implemented through noisy-score manipulations. Classifier-free guidance \citep{ho2022cfg} is often interpreted through powered conditional and unconditional densities, but the guided diffusion path need not match this pointwise interpretation \citep{chidambaram2024guidance,bradley2025cfg}. Complementary work attributes practical CFG failures to off-manifold guidance and modifies the sampling path accordingly \citep{chung2025cfgpp}, while recent work studies path consistency more directly through adaptive or fixed-point views of guidance \citep{wang2025goldencfg}. \citet{galashov2026guidance} learn time-dependent guidance rules by minimizing mismatch between the desired noised target and the guided diffusion process, including targets obtained by reward-tilting the clean distribution. Feynman--Kac correctors provide path-space corrections for annealing, guidance, and products of distributions \citep{skreta2025feynman}, while calibrated test-time guidance corrects biased noisy posterior guidance \citep{geyfman2026calibrated}. Together, these works emphasize that a transformation specified at the clean endpoint does not by itself determine its representation along the noisy path. We address this issue for global self-powering, deriving the induced noisy-score response directly from the denoising posterior rather than learning or correcting a guidance trajectory.

\paragraph{Heat flow, posterior moments, and negentropy.}

Our low-noise analysis builds on classical Gaussian heat-flow identities linking entropy, Fisher information, and derivatives of the evolving log density \citep{stam1959information,costa1985entropy,brigati2025heat}, together with classical Edgeworth/cumulant approximations to negentropy \citep{jones1987projection,comon1994ica}. Tweedie's formula relates the noisy score to the posterior mean \citep{efron2011tweedie}, while higher-order identities relate derivatives of the noisy log density to posterior moments \citep{meng2021higher,manor2024posterior,boys2024tweedie}.

A related inverse-problem and conditional-generation literature exploits this posterior geometry for sampling. Diffusion Posterior Sampling approximates posterior scores under noisy observations \citep{chung2023dps}, while filtering-based formulations provide asymptotically accurate posterior sampling for linear inverse problems \citep{dou2024filtering}. Subsequent methods explicitly exploit denoising-posterior covariance, including optimal-covariance and moment-matching approaches \citep{peng2024posteriorcov,rozet2024diffusionpriors}. \citet{rissanen2025freehunch} estimate denoising-posterior covariance using training statistics and trajectory curvature linked through second-order Tweedie identities, while \citet{ou2025optimalcov} use analytic covariance targets to improve probabilistic diffusion sampling. Most recently, \citet{mammadov2026exact} derive an exact posterior-score representation for linear Gaussian inverse problems with anisotropic noise. Separately, \citet{xi2026local} use Tweedie identities and posterior covariance to characterize quasi-locality of Gaussian-smoothed scores.

These works use posterior geometry for conditional inference, sampling, or score locality. Our object is instead the response obtained when the clean law itself is power tempered before diffusion. In particular, we characterize how denoising-posterior covariance enters this response and isolate residual contributions from covariance mismatch and non-Gaussian posterior shape. Our pointwise posterior-negentropy evolution further supplies the spatial-derivative control needed for the resulting low-noise score decomposition.

\paragraph{Synthetic self-training and negative extrapolation.}

Self-training generative models on their own samples can amplify distributional bias and lead to progressive degradation \citep{alemohammad2024mad}, motivating methods that stabilize or exploit synthetic-training dynamics \citep{gillman2024selfcorrecting}. NEON \citep{alemohammad2026neon} turns such degradation into a useful signal by fine-tuning on self-generated samples and extrapolating in the opposite parameter direction. As an application of our proposed theory, we show that under a locally power-tempered synthetic law, the propagated posterior-dependent field determines the perturbation seen by DSM and its pullback to parameter space, providing a diffusion-specific geometric interpretation of negative extrapolation.

\section{Connection to Negative Extrapolation}
\label{app:negative_extrapolation}

Negative extrapolation methods such as NEON \citep{alemohammad2026neon} form a synthetic-finetuned checkpoint $\theta_s$ and extrapolate away from it,
\begin{equation*}
\theta_{\rm NE}(w)=\theta_r-w(\theta_s-\theta_r).
\end{equation*}
For a sufficiently local synthetic gradient step,
\begin{equation*}
\theta_s-\theta_r
=\eta\delta\nabla_\theta V_{\rm full}(\theta_r)+O(\eta\delta^2+\eta^2),
\end{equation*}
so
\begin{equation*}
\theta_{\rm NE}(w)-\theta_r
=-w\eta\delta\nabla_\theta V_{\rm full}(\theta_r)
+O(\eta\delta^2+\eta^2).
\end{equation*}
Thus, under the power-tempering model, negative extrapolation reverses the local frozen-potential direction induced by synthetic training.  Approximating the propagated field by its posterior-covariance component, i.e., dropping the full residual $r_t=\nabla_x(\Phi_t+\nu_t)$, gives the potential $V_{\rm cov}$. This covariance potential can also be incorporated as a local regularizer around the reference checkpoint,
\begin{equation}
J_{\rm cov}(\theta)
=
R_{\rm anchor}(\theta)
+
\mu V_{\rm cov}(\theta),
\label{eq:Jcov}
\end{equation}
where $\mu \geq 0$ controls the regularization strength and  $R_{\rm anchor}$ is a standard DSM anchor and $V_{\rm cov}$ is evaluated with the frozen reference geometry. We use this objective only as a supporting test of whether the predicted covariance potential is useful beyond directional agreement.  Appendix~\ref{app:regularization} gives the corresponding $\epsilon$-parameterization and makes explicit the dependence on timestep sampling and noise weighting. A useful consequence of this score-space representation is that the covariance correction does not inherently require posterior sampling or explicit covariance estimation. The local field $B_t s_t$ can be represented directly through the reference score as
\[
B_t s_t
=
s_t+\sigma_t^2(\nabla_x s_t)s_t,
\]
and hence through a score-Jacobian--vector product. The posterior covariance therefore supplies the geometric interpretation of the correction, while its action is accessible directly from local differential information in the score field. Under an isotropic approximation to the posterior geometry, $B_t(x)\approx I$, the frozen covariance potential reduces to
\begin{equation*}
V_{\rm rad}(\theta)
=
\frac{1}{2}
\int
\rho(t)\,
\E_{p_{\theta_r,t}}
\left[
\|s_\theta(X,t)\|^2
\right]
\dd t,
\end{equation*}
which is ordinary score energy. Thus, within the local power-tempering model, score-norm regularization can be viewed as the isotropic limit of the covariance-weighted objective derived above. The distinction is geometric: the isotropic objective treats all score directions equally, whereas $V_{\mathrm{cov}}$ retains the direction-dependent weighting induced by denoising uncertainty. Because the corresponding field $B_t s_t$ admits the matrix-free score-Jacobian representation above, this anisotropic generalization does not conceptually require explicit posterior-covariance estimation. This provides a complementary motivation for score-norm regularization from current literature \citep{baptista2025memorization,niemann2026analysis}, while identifying the posterior geometry that is discarded by its isotropic approximation.

\end{document}